\documentclass[journal, onecolumn]{IEEEtran}
\usepackage{cite}
\usepackage{color}
\usepackage{amsfonts,amsmath,amssymb,amsthm}
\usepackage{graphicx}
\usepackage{epstopdf}
\usepackage{tikz, pgfplots,subcaption}
\usetikzlibrary{plotmarks,spy,backgrounds}
\pgfplotsset{compat=newest}
\definecolor{mycolor1}{rgb}{0.7,0,0}%
\definecolor{mycolor2}{rgb}{0.5,0.5,1}%
\definecolor{mycolor3}{rgb}{0,0,0.5}%
\usepackage{mathrsfs}

\definecolor{RED}{rgb}{0.7,0,0}
\definecolor{BLUE}{rgb}{0,0,0.69}
\definecolor{GREEN}{rgb}{0,0.6,0}
\definecolor{PURPLE}{rgb}{0.69,0,0.8}

\newtheorem{Theorem}{Theorem}
\newtheorem{Corollary}{Corollary}

\newcommand{\J}{{\mathbf{J}}}
\newcommand{\T}{{\mathbf{T}}}
\newcommand{\I}{{\mathbf{I}}}
\newcommand{\X}{{\mathbf{X}}}
\newcommand{\Y}{{\mathbf{Y}}}
\newcommand{\D}{{\mathbf{D}}}
\newcommand{\W}{{\mathbf{W}}}
\newcommand{\U}{{\mathbf{U}}}
\newcommand{\bb}{{\mathbf{b}}}
\newcommand{\h}{{\mathbf{h}}}
\newcommand{\x}{{\mathbf{x}}}

\newtheorem{Proof}{Proof}

\begin{document}
\title{Spectrum concentration in deep residual learning:  a free probability approach}
\author{Zenan Ling$^1$, Xing He$^1$, Robert C. Qiu$^{1,2}$,~\IEEEmembership{Fellow,~IEEE}
	\thanks{This work was partly supported by NSF of  China  No. 61571296 and (US)  NSF Grant No. CNS-1619250.

		$^1$ Department of Electrical Engineering,
		Center for Big Data and Artificial Intelligence, State Energy Smart Grid Research and Development Center, Shanghai Jiaotong
		University, Shanghai 200240, China. (e-mail: lingzenan@sjtu.edu.cn; rcqiu@sjtu.edu.cn;).
		
		$^2$ Department of Electrical and Computer Engineering,
		Tennessee Technological University, Cookeville, TN 38505, USA. (e-mail:rqiu@tntech.edu).
}}
\maketitle
\IEEEpeerreviewmaketitle
\begin{abstract}
  We revisit the weight initialization of deep residual networks  (ResNets) by introducing a novel analytical tool in free probability to the community of deep learning. This tool deals with the limiting spectral distribution of \textit{non-Hermitian} random matrices, rather than their conventional \textit{Hermitian} counterparts in the literature.  This new tool enables us to evaluate the singular value spectrum of the input-output Jacobian of a fully-connected deep ResNet in both linear and nonlinear cases. With the powerful tool of free probability, we conduct an asymptotic analysis of the (limiting) spectrum on the single-layer case, and then extend this analysis to the multi-layer case of an arbitrary number of layers.
  The asymptotic analysis illustrates the necessity and university of rescaling the classical random initialization by the number of residual units $L$, so that the  squared singular value of the associated Jacobian remains of order $O(1)$, when compared with the large width and depth of the network. We empirically demonstrate that the proposed initialization scheme learns at a speed of orders of magnitudes faster than the classical ones, and thus attests a strong practical relevance of this investigation.
\end{abstract}
\begin{IEEEkeywords}
	Residual network, weight initialization, random matrix theory, non-Hermitian free probability theory, Jacobian matrix, spectral density.
\end{IEEEkeywords}

\section{Introduction}

Deep neural networks have obtained impressive achievements in numerous fields from computer vision \cite{krizhevsky2012imagenet} to speech recognition \cite{mohamed2012acoustic} and natural language processing \cite{collobert2008unified}. Yet for all the successes won with these deep structures, we have gained only a rudimentary theoretical understanding of why and in what contexts they work well.
 Modern deep neural networks are typically trained with  gradient-based methods, where the (weight) initialization plays a crucial role in the efficient training of those deep models, as a result of the highly non-convex nature of the underlying objective function. Prior works~\cite{glorot2010understanding,He2015Delving,Pennington2017Resurrecting} have shown that, to prevent gradients from vanishing or exploding (which is believed to be the main difficulty in training deeper models that have more expressive power than shallower ones), one shall choose a proper initialization so that the deep network's  input-out Jacobian is well-conditioned. In other words, in order to preserve the norm of a randomly chosen error vector through backpropagation, the  squared singular values of the Jacobian matrix shall remain to be the order of \(O(1)\), compared with the (possibly) tremendous width or depth of the network. We refer to this property as the ``Spectrum Concentration'' of the Jacobian matrix, that  is different from the similar concept of ``Dynamical Isometry''~\cite{Saxe2013Exact} demanding that \textit{all} singular values remain close to \(1\).

\begin{figure*}
  \centering
  \includegraphics[width=1\textwidth]{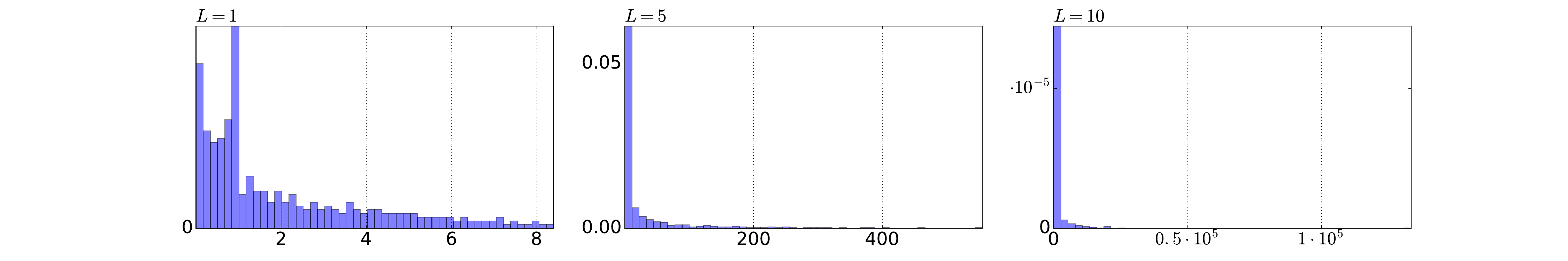}\\
  \caption{Empirical eigenvalue density of \(\J\J^\T\) with \(\J\) the input-output Jacobian of a fully connected (untrained) ResNet of width \(N=400\) and depth \(L=1,5,10\) with \(\sigma_w^2 = 2\).}\label{fig:intro}
\end{figure*}

 In particular, ResNet, as one of the most popular modern  deep network structures, has achieved the state-of-the-art performance on various challenging tasks~\cite{He2016Deep,He2016Identity}. Nonetheless, it is worthy noting that in practice the He initialization~\cite{He2015Delving} and the batch normalization (BN) technique~\cite{ioffe2015batch} are commonly combined together to ensure an effective training of ResNets.  Experiments in Fig.~\ref{fig:intro} show, on the other hand,  that the input-output Jacobian of a fully-connected ResNet (without BN) with He's initialization can be ill-conditioned, in the sense that most singular values are close to zero, while  many extremely large singular values lie in a heavy tail away from zero. This occurs even at the \textit{beginning} of the training procedure. Recall that, before the introduction of BN, various of deep networks have been successfully designed and trained without this catastrophic problem of exploding or vanishing gradients. This surprising empirical result
  naturally leads to the following question:
  \begin{center}
  \textit{Have we really used the ``good'' initialization for ResNets?}
  \end{center}

Among the commonly used random initialization schemes, the variances $\sigma_w^2$ of the Gaussian weights are \textit{always} normalized by the numbers of neurons of the corresponding layer (i.e., the \textit{width} $N$ of the network, for example in the case of He's initialization $\sigma_w^2 = \frac{2}{N}$). In contrast, the number of layers of the network (i.e., its \textit{depth} $L$), as another crucial parameter, has  been rarely taken into account.  In this article, exploiting advanced tools in random matrix theory in the regime of large network width and depth, we prove that, for ResNets, the variance of the random weights should also be scaled as a function of the number of layers, so as to prevent the gradients vanishing or exploding problem via spectrum concentration.

\subsection{Related work}
 The authors in \cite{Saxe2013Exact} start the consideration of the ill-conditioned Jacobian from random Gaussian initialization, and propose to use orthogonal weights initialization to achieve dynamical isometry in deep \textit{linear} networks. The recent works~\cite{Pennington2017Resurrecting,Pennington2018The} open the door for a direct application of random matrix theory, particularly \textit{free probability}, to  evaluating the Jacobian spectrum of a deep network, in which the singular value distribution of the Jacobian of a fully-connected network is analytically given as a function of  depth, random initialization and nonlinearity. In \cite{Hardt2016Identity} the authors prove the existence of a global optimal solution for linear ResNet, if the  spectral norms of the weights are bounded by \(O(1/L)\) and therefore, small random weights that is normalized by the layer number \(L\), helps deep residual learning. In \cite{Yang2017Mean}, the authors investigate the forward and backward signal propagation of  ResNet using mean field theory and discuss the  importance of the $O(1/L)$ scaling. However, the mean-field analysis only predicts the expectation of the Jacobian spectrum while the higher moments and the full distribution are not considered by the authors.  In \cite{Wojciech2017Deep}, the authors discuss the universal characters of the singular spectrum under the  $O(1/L)$ scaling with free probability. This work is related to ours but the derivation become tractable only by pre-assuming $\sigma_w^2=O(1/L)$ and  the necessity of the $O(1/L)$ scaling, and more general initialization settings, are not discussed in their work. Some similar results are presented in  \ref{sec:Full spectrum characterization} for completeness.
\subsection{Our contributions}
Based on recent advances in free probability theory, we establish a general framework for the spectral analysis of the input-output Jacobian of a ResNet, for Gaussian and orthogonal random initialization with various nonlinear activation functions.  The conditions for \emph{necessity} and \emph{university} of taking $\sigma_w^2 = O(1/L)$ is unified under the proposed analysis framework.

More concretely, we extend the framework established in \cite{Pennington2017Resurrecting} to a  non-Hermitian setting so as to overcome the (non-trivial) technical difficulty (mentioned in \ref{sec:singlelayer}) arising from studying the spectrum of the input-output Jacobian of a single layer ResNet. This result is then extended to the multi-layer case, for which we calculate the expectation and variance of the \textit{full} spectrum. The results of the expectation and variance demonstrates the necessity of taking \(\sigma_w^2 = O(1/L) \) to ensure the  aforementioned key property of spectrum concentration to facilitate training. Furthermore, the full spectrum characterization in the case of $\sigma_w^2 = O(1/L)$ is provided. The result illustrates that it suffices to take \(\sigma_w^2 = O(1/L) \) to ensure the  squared singular values of the  aforementioned Jacobian to be of order \(O(1)\), for both  random Gaussian and orthogonal weights with \emph{any} nonlinearity,  which meets some mild assumption.  The theoretical results are  corroborated by empirical evidences on popular CIFAR-10 dataset~\cite{cifar}. For the sake of simplicity, some detailed proofs and complementary experiments are deferred to Appendix.
\section{Problem Statement and Preliminaries}
\label{sec:problem-preliminary}
\subsection{Problem set up}
\label{sec:problem-setup}
Denote the output  vector of the \((l-1)\)-th layer \(\x^l \in \mathbb{R}^N\), weight matrix \(\W^l \in \mathbb{R}^{N \times N}\), bias vector \(\bb^l\in \mathbb{R}^N\) of the layer \(l\) and pre-activation $\h^l$, then the forward dynamics of  a fully connected ResNet
 without BN of depth $L$ is given by:
\begin{equation}
\begin{split}
    \h^l &= \W^l \x^{l-1} + \bb^{l}, \\
    \x^l &= \x^{l-1} + \phi(\h^{l}), \quad  \mbox{for} \quad l=1,\cdots,L,
\end{split}
\label{resnetdynamic}
\end{equation}
where \(\x_0 \in \mathbb{R}^N\) is the input data of the network and \(\phi:\mathbb{R}\mapsto\mathbb{R}\) denotes the pointwise nonlinearity. The associated input-output Jacobian is given by
\begin{equation}
  \J = \frac{\partial \x^L}{\partial \x^0}=\prod_{l=1}^L (\I_N + \D^l\W^l ).
\end{equation}
 with diagonal \(\D^l\) such that \(\D^l_{ii} = \phi^\prime(\x_i^{l-1})\).

We are interested in the initial state of the training procedure of a ResNet  described in \eqref{resnetdynamic} by considering
two popular random weight initializations:  random Gaussian weights with \(\W_{ij}^l \sim \mathcal{N}(0,\sigma_w^2/N)\), and
 random orthogonal weights that satisfies \(\W^l (\W^l)^\T=\sigma_w^2\I_N\).

For nonlinearity, we make two wild assumptions  that $\int {\phi^{\prime}(x)^{2}\mathcal{D}x}\neq  0$ and $\int { \phi(x)\mathcal{D}x} $ is non-negative, where \( \mathcal{D}x = e^{-\frac{x^2}{2}}dx / \sqrt{2\pi}\) denotes the standard Gaussian measure.  Note that almost all of the frequently-used activation functions meet these  conditions.

Moreover, following several related work~\cite{Pennington2017Resurrecting,Schoenholz2017Deep,Yang2017Mean}, we make a \emph{key assumption} that  the weights in the forward and backward propagation are independent. Mathematically, this assumption is incorrect because the activations of deeper layers depend explicitly on the weight matrices of shallower layers. However, theoretical computations become tractable under this assumption and  the empirical results show a strong support for it. There must be a phase transition that this as-
sumption breaks down after some training steps.  Quantitatively controlling this approximation may be  quite complicated and we leave this investigation to the future work.
\subsection{Signal propagation}
\label{subsec:signal-propagation}
 For large \(N\), the empirical distribution of $\h_i^{l}$ converges to a zero mean Gaussian since that each $\h^l=\W^l \x^{l-1}+\bb^l$ is a weighted sum of a large number of uncorrelated random variable, i.e., the weights and biases which are independent of the activation in previous layers. Let \(q^l \equiv \frac{1}{N}\sum\nolimits_{i=1}^{N}(\h_i^l)^2\) denotes the variance of the pre-activation \(\h^l\). For ResNet \eqref{resnetdynamic}, the  recursive equation for \(q^l\) is given by,
\begin{equation}
\begin{split}
q^{l+1} = &q^l + \sigma_w^2 \int { \phi^2\left(\sqrt{q^l}x\right)\mathcal{D}x}\\
&+2\sigma_w^2 \left[\sum\limits_{k=0}^{l-1}\int { \phi \left(\sqrt{q^{k}}x \right)\mathcal{D}x}\right]\int {\phi \left(\sqrt{q^l}x\right)\mathcal{D}x },
\end{split}
\label{recursiveeq}
\end{equation}
with initial condition $q^0 =\frac{1}{N}\sum\nolimits_{i=1}^{N}(\h_i^0)^2 $.

The detailed derivation for the equivalent argument is provided in \cite{Wojciech2017Deep,Yang2017Mean}. The recursion relation \eqref{recursiveeq} for ResNets is essentially different from that of a fully-connected vanilla neural network without residual connection  that the biases have no influence.  Moreover, one can easily observe that \(q^{l+1}\) is a result of adding positive terms to the previous \(q^l\). Thus, the variance of pre-activations grows with the depth and  \emph{no non-trivial fixed point exits} in the recursion \eqref{recursiveeq}.
\subsection{Hermitian free probability theory}
Free probability generalizes probability theory to algebras of non-commutative random variables, which is notably the case of the algebra of random matrices~\cite{Nica2006Lectures,Tao2012Topics}.
When a pair of random matrices is free, the eigenvalue distribution of their combinations (sum, product, etc.) can then be determined through specific analytical tools, introduced next.\footnote{ In the section that follows, the argument \(z\) will be frequently dropped for notational simplicity.   \(f^{-1}\) denotes the functional inverse of \(f\).}

The spectral density of a random Hermitian matrix \(\X \in \mathbb{R}^{N\times N}\) is defined as
\(\rho_{\X}(\lambda)=\frac{1}{N}\sum\nolimits_{k=1}^{N} {\delta(\lambda-\lambda_k(\X))}\),
where \(\lambda_k(\X)\) (\(k=1,\cdots,N\)) denote the \( N \) eigenvalues of \(\X\). The limiting spectral density is defined as the limit of \(\rho_{\X}(\lambda)\) as \(N \to \infty  \), if it exits.

The Stieltjes transform of \(\rho_\X\) is defined as
\begin{equation}
G_{\X}(z)\equiv\int_\mathbb{R} {\frac{{{\rho _\X}(t)}}{{z - t}}} dt,
\label{eq:stj}
\end{equation}
where \(z \in \{z:z \in \mathbb{C}, \Im(z)>0\}\).
The spectral density can be recovered from the Stieltjes transform using the inversion formula,
\begin{equation}
\rho_\X(\lambda)=-\frac{1}{\pi}{\lim_{\epsilon \to 0^+}}\Im G_{\X}(\lambda+i\epsilon).
\label{inversion formula}
\end{equation}
The Stieltjes transform can be typically expanded into a power series  as \begin{equation}
G_\X(z)= \sum\limits_{k=0}^{\infty} {\frac{m_k}{z^{k}}},
\end{equation}
with the matrix moments
\begin{equation}
m_k=\int { \rho(\lambda)\lambda^k}d\lambda,
\end{equation}
which further determine the moment generating function \(M_\X\) (also referred to as the M-transform) of the random matrix X,
\begin{equation}
M_\X(z)=zG_\X(z)-1= \sum\limits_{k=1}^{\infty} {\frac{m_k}{z^{k}}}.
\label{M2G}
\end{equation}
And the S-transform of \(\X\) is defined as,
\begin{equation}
\label{s-transform}
S_\X(z)\equiv\frac{1+z}{zM^{-1}_{\X}(z)}.
\end{equation}
The power series \(G_\X(z)\) can be inverted (for composition of formal power series), in the form,
\begin{equation}
G^{-1}_\X(z) = \frac{1}{z} + \frac{1}{z}\sum\limits_{k=1}^{\infty}r_{k}z^{k}=\frac{1}{z} + R_\X(z).
\label{R-transform}
\end{equation}
The power series \(R_\X(z)\) is called the R-transform of \(\X\) and its coefficients are called the free cumulants.
For any two freely independent non-commutative random variables \(\X,\Y\), the R- and S-transform have the following definite (convolution) properties,
\begin{equation}
R_{\X+\Y}(z)=R_\X(z)+R_\Y(z),
\label{Rdefinition}
\end{equation}
\begin{equation}
S_{\X\Y}(z)=S_\X(z)S_\Y(z).
\label{Sdefinition}
\end{equation}
As such, the R-transform\textit{ linearizes} free additive convolution and the S-transform of the matrix multiplication  \(\mathbf{XY}\) is simply the multiplication of their S-transforms.

Moreover,  R- and S-transforms relate through~\cite{Nica2006Lectures},
\begin{equation}
S_\X(zR_\X(z))=\frac{1}{R_\X(z)}.
\label{S2R}
\end{equation}
\subsection{Non-Hermitian free probability theory}
Consider the single layer case in our problem. Let \(\J_l:=\I + \D^l\W^l\) denotes the input-output Jacobian matrix of the layer \(l\). Expand \(\J_l\J_l^\T\) and we have that
\begin{equation}
\J_l\J_l^\T = \I + \D^l\W^l(\W^l)^\T\D^l + \D^l\W^l + (\W^l)^\T\D^l.
\label{jjexpand}
\end{equation}

Note that the objective of interest \(\J_l\J_l^\T\) is \emph{Hermitian} but the resulting four terms of expansion are \emph{not freely independent} and thus can not be handled with a single R-transform. On the other hand, the term in \(\J_l\) are \textit{free} but they are \textit{non-Hermitian}. As such, we perform an extension of conventional (Hermitian) free probability to non-Hermitian random matrices~\cite{Cakmak2012Non} to evaluate the limiting eigenvalue distribution of \(\J_l\J_l^\T\).

Consider a Hermitian matrix \(\widetilde{\X}\) such that the eigenvalue distribution of \(\widetilde{\X}\) is
\begin{equation}
\rho_{\widetilde{\X}}(\lambda)=\frac{\rho_{\sqrt{\X\X^\T}}(\lambda)+\rho_{\sqrt{\X\X^\T}}(-\lambda)}{2}
\end{equation}
where \(\widetilde{\X}\) is symmetrized singular value version of \(\X\).
The following equation establishes the connection between  \(\widetilde{\X}\) and \(\X\X^\T\),
\begin{equation}
G_{\widetilde{\X}}(z)=zG_{\X\X^\T}(z^2),
\label{G2Gtilde}
\end{equation}

A random matrix \(\X\) is called \emph{R-diagonal} if it can be decomposed as \(\mathbf{X=UY}\), such that \(\U\) is Haar unitary and free of \(\Y=\sqrt{\X\X^\T}\).
If the  free random matrices \(\X\) and \(\Y\) are R-diagonal, then we have,
\begin{equation}
R_{\widetilde{\mathbf{X+Y}}}=R_{\widetilde{\X}}+R_{\widetilde{\Y}}.
\label{rdiagR}
\end{equation}
where $R_{\widetilde{\mathbf{X}}}(z) = \sum \nolimits_{k=1}^{\infty} r_{2k-1}z^{k} $ generates the cumulants $r_{2k-1}$.

For a random matrix \(\X\), the S-transforms of \(\widetilde{\X}\) and \(\X\X^\T\) have the following relation:
\begin{equation}
S_{\widetilde{\X}}(z)=\sqrt{\frac{z+1}{z}S_{\X\X^\T}(z)}.
\label{S4xxxt}
\end{equation}

\section{Theoretical Results}
\label{sec:theory}
Equipped with the aforementioned free probability tool, we are in the position to study the asymptotic spectrum of the Jacobian matrix in the simultaneously large \(N,L\) limit. As mentioned in Section~\ref{subsec:signal-propagation}, \emph{no non-trivial fixed point exits} in the recursion \eqref{recursiveeq} and we can not simply assume that  $\D^l$ equals to each other as in the case of vanilla fully connected networks~\cite{Pennington2017Resurrecting,Pennington2018The}. Thus, we provide the analysis for the single layer first and extend the result to the multi layer case through  S-transform and power series expansion. The necessity of taking $\sigma_w^2=O(1/L)$ is proved by investigating the expectation and variance of the spectral density distribution. Finally,  the university of taking $\sigma_w^2=O(1/L)$ is discussed  by  investigating the full spectrum characterization.
\subsection{Single layer case}
\label{sec:singlelayer}
First, we deduce the  equation for solving the Stieltjes transform $G_{\J_l\J_l^\T}$ of ${\J_l\J_l^\T}$.
According to~\cite{Cakmak2012Non}, we have
\begin{equation}
G_{\J_l\J_l^\T} = G_{(\I+\D^l\W^l)(\I+\D^l\W^l)^\T} = G_{(\U+\D^l\W^l)(\U+\D^l\W^l)^\T},
\end{equation}
where \(\mathbf U\) is a random Haar unitary matrix and free of \(\W^l\D^l\).
Note that \(\widetilde{\U}\) and  \(\widetilde{\D^l\W^l}\) are R-diagonal, with \eqref{rdiagR} we have,
\begin{equation}
R_{\widetilde{\U+\D^l\W^l}}(z)=R_{\widetilde{\U}}(z)+R_{\widetilde{\D^l\W^l}}(z).
\end{equation}
According to the definition of R-transform \eqref{R-transform}, we have,
\begin{equation}
\begin{split}
z=&R_{\widetilde{\U}}\left[G_{\widetilde{\U+\D^l\W^l}}(z)\right]+
R_{\widetilde{\D^l\W^l}}\left[G_{\widetilde{\U+\D^l\W^l}}(z)\right]\\
&+\frac{1}{G_{\widetilde{\U+\D^l\W^l}}(z)}
\label{G2RUDW}
\end{split}
\end{equation}
With \eqref{G2Gtilde}, we have,
\begin{equation}
G_{\J_l\J_l^\T}=G_{(\U+\D^l\W^l)(\U+\D^l\W^l)^\T}(z) = \frac{1}{\sqrt z}G_{\widetilde{\U+\D^l\W^l}}(\sqrt z).
\label{G2GtildeUDW}
\end{equation}
By substitute \eqref{G2GtildeUDW} to \eqref{G2RUDW}, we have the  following theorem.
\begin{Theorem}[Single layer case]
\normalfont For all $z \in \mathbb{C}$ with positive imaginary part, denote \(G_{\J_l\J_l^\T}(z)\) the (limiting) Stieltjes transform of \(\J_l\J_l^\T\). Then, as $N \to \infty$, we have
\begin{equation}
\begin{split}
&\sqrt{z} G_{\J_l\J_l^\T}(z)  \left[R_{\widetilde{\U}}\left(\sqrt{z} G_{\J_l\J_l^\T}(z)\right) + R_{\widetilde{\W^l\D^l}}\left(\sqrt{z} G_{\J_l\J_l^\T}(z)\right)\right]\\
&=zG_{\J_l\J_l^\T}(z) -1 ,
\end{split}
\label{GJJeq}
\end{equation}
where  \(\mathbf U\) is a random Haar unitary matrix and free of \(\D^l\W^l\). The correct root is selected by the asymptotic behavior \( G_{\J_l\J_l^\T}(z) \sim \frac{1}{z}\) as \(z \to \infty\)~\cite{Tao2012Topics}.
\label{Master equation}
\end{Theorem}

 Based on Theorem.~\ref{Master equation},  the detailed procedure for calculating the density of $\J_l\J_l^\T$ is summarized as follows:
\begin{enumerate}
\item Calculate \(M_{{\D^l}^2}(z)\) with \eqref{M2G}, then calculate \(S_{{\D^l}^2}(z)\) with \eqref{s-transform};
\item Calculate \(S_{{\D^l\W^l}({\D^l\W^l})^\T}(z)\) with \eqref{Sdefinition};
\item Calculate \(S_{\widetilde{\U}}(z)\) and \(S_{\widetilde{\D^l\W^l}}(z)\) with \eqref{S4xxxt};
\item Calculate \(R_{\widetilde{\U}}(z)\) and \(R_{\widetilde{\D^l\W^l}}(z)\) with \eqref{S2R};
\item Calculate \(G_{\J_l\J_l^\T}(z)\) with \eqref{GJJeq};
\item Calculate  the spectral density with \eqref{inversion formula}.
\end{enumerate}
 The calculation of \(S_{{\D^l\W^l}({\D^l\W^l})^\T}(z)\) in step 2 requires the information of  \( S_{\W^l(\W^l)^\T}(z)\) and \( M_{\D^l}^2(z)\). For scaled Gaussian weights, the spectral density distribution follows the famous \emph{Marchenko-Pastur Law} (M-P Law)~\cite{V1967Distribution}, \begin{equation}
 \rho_{\W^l(\W^l)^\T}(\lambda) =\frac{\sqrt{(4\sigma_w^2-\lambda)\lambda}}{2\pi\sigma_w^2},
 \end{equation}
 for $\lambda \in [0, 4\sigma_w^2]$.
 Through \eqref{eq:stj}, \eqref{M2G} and \eqref{Sdefinition}, it is easy to deduce that,
 \begin{equation}
S_{\W^l(\W^l)^\T}(z) =\frac{1}{\sigma_w^{2}(1+z)}.
\label{s_guass}
 \end{equation}
 For scaled orthogonal weights, it is obvious that
\begin{equation}
\rho_{\W^l(\W^l)^\T}(\lambda) = \delta(\lambda-\sigma_w^2).
\end{equation}
Using \eqref{eq:stj}, \eqref{M2G} and \eqref{Sdefinition} again, we obtain \( S_{\W^l(\W^l)^\T}(z)=1\).

Moreover,   \(\D^l\) is a diagonal matrix with  \(\D^l_{ii} = \phi^\prime(\h_i^{l-1})\), so $\phi^\prime(\h_i^{l-1})^2$ is the eigenvalue of \((\D^l)^2\). The empirical distribution of pre-activations $\h_i^l$ converges to a Gaussian with zero mean and variance $q^l$, in
the large N limit. Therefore, for any nonlinearity $\phi(x)$, we have, through \eqref{eq:stj} and \eqref{M2G},
\begin{equation}
M_{\D^l}^2(z) = \int \frac{\phi^{\prime}(\sqrt{q^l}x)^2}{z-\phi^{\prime}(\sqrt{q^l}x)^2}\mathcal{D}x.
\end{equation}

\begin{figure*}
  \centering
  \includegraphics[width=1\textwidth]{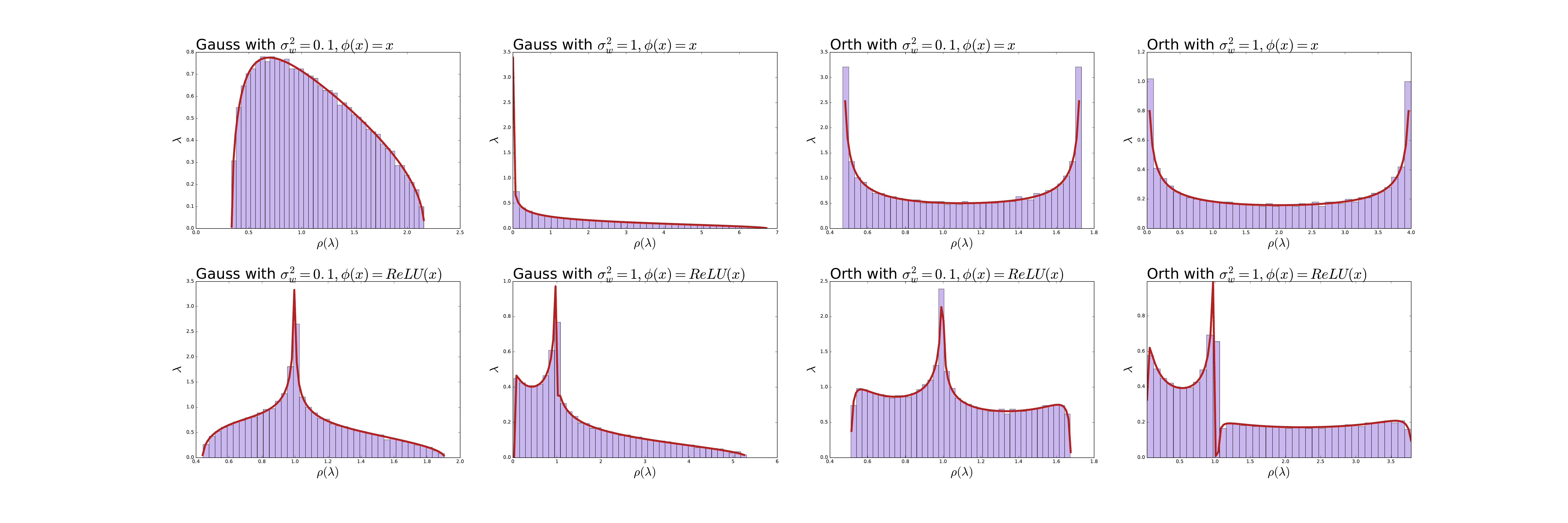}
  \caption{Empirical eigenvalue density (purple) and limiting distribution (red)   procedure of \(\J_l\J_l^\T\) for Gaussian  and orthogonal weights with \(N=400\),  $\sigma^2_w = 0.1$ and $1$.}\label{fig:gauss-orth-spectrum-single-layer}
\end{figure*}
In Fig.~\ref{fig:gauss-orth-spectrum-single-layer}, we plot the empirical eigenvalue density of \(\J_l \J_l^\T \) (in purple) and the limiting distribution (in red) calculated from the above procedure, for Gaussian and orthogonal weights with \( \sigma_w^2 = 0.1 \) or \(1\), \(\phi(x)=ReLU(x)\) or $x$.

In principle, this procedure can be carried out  for an arbitrary choice of nonlinearity so that we can deduce the limiting spectral distribution for $\J_l\J_l^\T$ for any $\phi(\cdot)$. However, the solution of \eqref{GJJeq} can be really complicated and unenlightening, and thus the calculation for multi-layer case can not be carried out. Inspired by \cite{Pennington2018The}, we  investigate the lower moments,  i.e., the expectation \(\mu_{\J_l\J_l^\T}\) and variance \(\sigma^2_{\J_l\J_l^\T}\), of the limiting eigenvalue density of \(\J_l\J_l^\T\) instead.

Note that the cumulants of \(R_{\widetilde{\D^l\W^l}}(z)\) can be calculated in terms  of the series expansions of \( S_{\W^l(\W^l)^\T}(z)\) and  \(M_{{\D^l}^2}(z)\), which are defined as,
\begin{equation}
 S_{\W^l(\W^l)^\T}(z) \equiv \sigma_w^{-2} (1+ \sum \nolimits_{k=1}^{\infty}s_kz^k),
 \label{s_expending}
\end{equation}
\begin{equation}
M_{{\D^l}^2}(z) \equiv \sum\nolimits_{k=1}^{\infty} {d^{(l)}_k}{z^{-k}},
\label{d_expanding}
\end{equation}
  where the moments of \({\D^l}^2\) are given by
\begin{equation}
d^{(l)}_k = \int {\left[\phi^{\prime}(\sqrt{q^l}x)\right]^{2k}\mathcal{D}x}.
\end{equation}
Moreover, both the Stieltjes and R-transtrom  in \eqref{GJJeq} can be expanded into power series. Thus, we can obtain the low order moments by expanding \eqref{GJJeq}. After a tedious manipulation, we  get the first and second order moment of the spectral density of \(\J_l \J_l^\T \),
\begin{equation}
\begin{split}
m_{1}^{(l)} &= 1 + \sigma_w^2d^{(l)}_1 ,\\
m_{2}^{(l)} &= 1+ \sigma_w^2\left(4d^{(l)}_1+\sigma_w^2\left(d^{(l)}_2 -(d^{(l)}_1)^2s_1 \right)\right).
\end{split}
\end{equation}
As $\mu_{\J_l\J_l^\T} = m_{1}^{(l)}$ and $\sigma^2_{\J_l\J_l^\T} =m_{2}^{(l)} -  (m_{1}^{(l)})^2 $, we have the following corollary.
\begin{Corollary}
\normalfont Assuming that \(d^{(l)}_1 \neq 0\), the expectation \(\mu_{\J_l\J_l^\T}\) and variance \(\sigma^2_{\J_l\J_l^\T}\) of the limiting eigenvalue density of \(\J_l\J_l^\T\) are given by,
\begin{equation}
\begin{split}
\label{singlemuvar}
\mu_{\J_l\J_l^\T} &= 1 + \sigma_w^2d^{(l)}_1 ,\\
\sigma^2_{\J_l\J_l^\T} &=  \sigma_w^2\left(2d^{(l)}_1+ \sigma_w^2 \left(d^{(l)}_2  - (d^{(l)}_1)^2 \left(1+s_1\right)\right) \right),
\end{split}
\end{equation}
where  $s_1$ and $d_1$, $d_2$ are defined in \eqref{s_expending} and \eqref{d_expanding}, respectively.
\label{th:singlemuvar}
\end{Corollary}



\subsection{Extension to the multi-layer case}
\label{subsec:multi-layer-case}
We use the important property of the S-transform \eqref{Sdefinition} to extend the results of single layer to multi-layer case.
Since the trace operator is \emph{cyclic-invariant}, we have
\begin{equation}
\begin{split}
S_{\J\J^\T}=S_{\prod\limits_{l = 1}^L {(\I + {\W^l}{\D^l}){{(\I + {\W^l}{\D^l}})^\T}} }
                 =S_{\prod\limits_{l = 1}^L {\J_l\J_l^\T} }=\prod\limits_{l = 1}^LS_{ {\J_l\J_l^\T} }.
\end{split}
\label{s*s}
\end{equation}

We see that the S-transform of \(\J\J^\T\) is simply given by the product of the S-transform of each residual unit \(\J_l \J_l^\T\).
Built upon this observation, the expectation \(\mu_{\J \J^\T}\) and variance \(\sigma^2_{\J \J^\T}\) of the limiting eigenvalue density of \(\J\J^\T\) are given as follows,
\begin{Theorem}[Mean and Variance of the Limiting Spectrum Density]
\label{thm:multi-layer-case}
\normalfont For \( \J = \prod_{l=1}^L (\J^l) \),  the mean and variance of the limiting eigenvalue density of ${\J\J^\T}$ are given by
\begin{equation}
\mu_{\J\J^\T}=\prod_{l=1}^{L}{\mu_{\J_l\J_l^\T}}, \quad
\sigma^2_{\J\J^\T}=\left(\prod_{l=1}^{L}{\mu_{\J_l\J_l^\T}}\right)^2\sum_{l=1}^{L} \frac{\sigma^2_{\J_l\J_l^\T}}{\mu_{\J_l\J_l^\T}^2}.
\end{equation}
as $N \to \infty$.
\end{Theorem}
We refer the readers to Proof~\ref{proof:Multi-layer} in Appendix  for detailed deduction. Using Theorem~\ref{thm:multi-layer-case}, the expectation and variance  of the limiting eigenvalue density of \(\J\J^\T\) can be diretly computed with the results in Corollary 1 of the single layer case \eqref{singlemuvar}.
To ensure the mean squared singular value of the input-output Jacobian to be of order $O(1)$ for large \(L\), we shall have \(\mu_{\mathbf{J}\mathbf{J}{^\T}} = O(1)\). This order requirement further indicates that, for both Gaussian and orthogonal weights, we shall have
\begin{equation}
\prod_{l=1}^{L} (1 + d^{(l)}_1 \sigma_w^2) = O(1),
\end{equation}
using Corollary~\ref{th:singlemuvar}.

Since that setting \(d^{(l)}_1 \to 0 \), or  equivalently \(\phi^{\prime}(z) \to 0\),   implies  that almost \textit{all} neurons are inactivated and will lead to a total failure of training, it is \emph{necessary} to scale the weight variance  with the layer number,
\begin{equation}
\sigma_w^2 = O(\frac{1}{L}),
\end{equation}
to ensure that \(\mu_{\mathbf{J}\mathbf{J}{^\T}} = O(1)\) for large \(L\).

 For the vanilla fully connected network, the variance $\sigma^2_{\J\J^\T}$ may still grow in an unbounded way with the layer number $L$ even if  $\mu_{\J\J^\T}=O(1)$~\cite{Pennington2017Resurrecting,Pennington2018The}. Only orthogonal initialization can yield a stable Jacobian spectral distribution for any choice of nonlinearity with $\phi^{\prime}(0)=1$. However, for deep ResNets, one can easily observe that the variance of  the squared singular values of the input-output Jacobian is of order \(O(1)\) if \(\sigma_w^2 = O(1/L)\). Therefore, we have the following corollary.
\begin{Corollary}
	\normalfont  For the ResNet which is defined as \eqref{resnetdynamic}, it is necessary to take \(\sigma_w^2 = O(1/L)\) to ensure that \(\mu_{{\J}{\J}{^\T}} = O(1)\) and \(\sigma^2_{{\J}{\J}{^\T}} = O(1)\), as $L \to \infty$.
	\label{th:spec-concentrated}
\end{Corollary}
 This observation illustrates a \emph{universality}  in the  Jacobian spectrum of the deep ResNet.  In particular, for both  scaled Gaussian and orthogonal weights, it is necessary to take \( \sigma_w^2 = O \left( 1/L \right) \) to ensure that not only the expectation but also the variance of the squared singular values of the  Jacobian matrix to be of order \(O(1)\) with \emph{any} nonlinearity, that meets the assumption made in~\ref{sec:problem-setup}.

\subsection{Full spectrum characterization}
\label{sec:Full spectrum characterization}
We have proved that setting $\sigma_w^2=O(1/L)$ is necessary to  keep the order of the expectation and variance of the Jacobian spectrum of ResNet, in the large $L$ limit.  We discuss the full characterization of the input-output Jacobian spectrum in this subsection. Fortunately, letting \(\sigma_w^2 = O(1/L)\) makes the deduction of the spectral density distribution of $\J\J^l$ tractable.
Assuming that $\sigma_w^2 = c/L$, where \(c\) is a positive constant of  order one. Then it is easy to obtain that as $L\rightarrow \infty$,
\begin{equation}
R_{\widetilde{\D^l\W^l}}(z) = \sum \nolimits_{k=1}^{\infty} r_{2k-1}z^{k} = \frac{cd^{l}_1}{L}z + O(\frac{1}{L^2}).
\end{equation}
which leads to
\begin{equation}
R_{\widetilde\J_l} = R_{\widetilde{\U + \D^l\W^l}}(z) = 1 +\frac{ cd^{l}_1}{L}z + O(\frac{1}{L^2}).
\end{equation}
Here, $r_1 = \frac{cd^{l}_1}{L}$ donates the mean squared radius of $\D^l\W^l$.

Solving  $S_{\widetilde\J_l}$ with \eqref{S2R} and substituting  $S_{\widetilde\J_l}$ to \eqref{S4xxxt}, we have,
\begin{equation}
S_{\J_l\J_l^{\T}} = 1-\frac{cd^{(l)}_1}{L}(2z+1)+O(\frac{1}{L^2}).
\end{equation}
Taking the logarithm of \eqref{s*s} yields,
\begin{equation}
\begin{split}
\ln S_{\J\J^{\T}}(z) &= \sum \limits_{l=1}^{L} \ln \left(1-\frac{cd^{(l)}_1}{L}(2z+1)\right)+ O(\frac{1}{L^2})\\
& \approx -\frac{c}{L} \sum \limits_{l=1}^{L}d^{(l)}_{1}(2z+1)\\
&= -\theta(2z+1),
\end{split}
\end{equation}
where \(\theta \equiv \frac{c}{L}\sum\limits_{l=1}^{L}d_{1}^{(l)}\).
Then, we obtain the S-transform of $\J\J^{\T}$,
\begin{equation}
S_{\J\J^{\T}}(z) = e^{-\theta(2z+1)}.
\label{SL}
\end{equation}
According to \eqref{s-transform}, we have,
\begin{equation}
S_{\J\J^{\T}}(z)=\frac{1+z}{zM^{-1}_{{\J\J^{\T}}}(z)}.
\end{equation}
Substituting $z \rightarrow M_{{\J\J^{\T}}}(z)$ yields,
\begin{equation}
S_{\J\J^{\T}}(M_{{\J\J^{\T}}}(z))=\frac{1+M_{{\J\J^{\T}}}(z)}{zM_{{\J\J^{\T}}}(z)}.
\end{equation}
According to  \eqref{M2G}, $M_{{\J\J^{\T}}}(z) = zG_{{\J\J^{\T}}}(z)-1$. Thus,
\begin{equation}
S_{\J\J^{\T}}(zG_{\J\J^{\T}}(z)-1)= \frac{G_{\J\J^{\T}}(z)}{zG_{\J\J^{\T}}(z)-1}.
\label{S2GJJ}
\end{equation}
Substituting \eqref{SL} to \eqref{S2GJJ},  we finally get the equation of the Stieltjes transform  \(G_{\J\J^\T}(z)\) as the  following corollary.

\begin{Corollary}
\label{th:Full-spectrum}
\normalfont Taking \(\sigma_w^2=c/L\), where \(c\) is a positive constant, for both Gaussian and orthogonal weights, the   Stieltjes transform \(G_{\J\J^\T}(z)\) satisfies,
\begin{equation}
G_{{\J}{\J}{^\T}}(z)e^{\theta \left(2zG_{{\J}{\J}{^\T}}(z) - 1\right)} =  zG_{{\J}{\J}{^\T}}(z)-1,
\label{stjloc/L}
\end{equation}
where we define \(\theta \equiv \frac{c}{L}\sum\limits_{l=1}^{L}d_{1}^{(l)}\).
\end{Corollary}

A recent work~\cite{Wojciech2017Deep} also obtains the similar result as in Corollary~\ref{th:Full-spectrum}. Different from our general case, their work makes the explicit assumption that \(\sigma_w^2 = O(1/L)\).

Next, the detailed but brief deduction of the condition number of $\J$ is provided here. The condition number is defined as the ratio of the maximal  and  minimal singular values of $\J.$  It measures   the stability of the spectrum.
 For the deduction of the condition number $cond(\J)$ of $\J$, we use a trick~\cite{Akemann2013Products} by multiplying \(z\) on the both sides of \eqref{stjloc/L},
\begin{equation}
zG_{{\J}{\J}{^\T}}(z)e^{\theta \left(2zG_{{\J}{\J}{^\T}}(z) - 1\right)} = z\left( zG_{{\J}{\J}{^\T}}(z)-1 \right).
\label{stjloc/L*z}
\end{equation}
Note that  \(\frac{{dz}}{{dG}} = 0\) at the endpoints of support of the  spectrum ~\cite{Akemann2013Products}. By differentiating both sides of \eqref{stjloc/L*z}, we have
\begin{equation}
e^{\theta \left(2zG_{{\J}{\J}{^\T}}(z)-1\right)}\left(2\theta zG_{{\J}{\J}{^\T}}(z) + 1\right)=z
\label{dstjloc/L*z}
\end{equation}
Substitute \eqref{dstjloc/L*z} to \eqref{stjloc/L}  and the final result gives that
\begin{equation}
\lambda_{\pm} = \left(1 + \theta \pm \sqrt{\theta^2+2\theta}\right)e^{\pm\sqrt{\theta^2+2\theta}},
\end{equation}
where \(\lambda_{\pm}\) donate the maximal  and  minimal eigenvalue of ${\J}{\J}{^\T}$ respectively. Thus, the conditional number of the input-output Jacobian matrix $\J$ is,
\begin{equation}
cond(\J)=\sqrt{\frac{\lambda_{+}}{\lambda_{-}}}=\left(1 + \theta + \sqrt{\theta^2+2\theta}\right)e^{\sqrt{\theta^2+2\theta}}.
\end{equation}

\section{Experiments}
In this section, we provide empirical evidence to validate the theoretical results in Section~\ref{sec:theory}. Experiments on fully-connected and convolutional ResNets are performed on CIFAR-10.
The standard CIFAR-10 datasets augmented with random flips and crops, and random saturation, brightness, and contrast perturbations are applied.
Two commonly used optimizers: SGD-Momentum and ADAM~\cite{Kingma2014Adam} are adopted.  The observation that two different methods: SGD-Momentum and ADAM  give very similar results, indicates the robustness of our approach. See Section~\ref{sm-sec:simu} in Appendix for the results of ADAM. Ten repeated experiments are conducted for each setting and the average results are reported  here.
\label{sec:experiments}
\subsection{Fully-connected ResNet}
In this case, the input dimension is reduced to \(N=400\) with a fully-connected layer of size \(1\,728\times 400\). We train a fully-connected ResNet\footnote{ In fully-connected ResNet case, every residual unit contains a single layer without batch normalization.} of depth \(L=100\) and width \(N=400\) for \(200\) epoches with a mini-batch size of \(128\). Four initialization scalings: \(\sigma_w^2= 1,c/L\), ($c=1$, $0.1$, $0.01$), are explored here.

\subsubsection{Jacobian spectrum at initialization} 
\begin{figure}[htbp]
  \centering
  \includegraphics[width=1\textwidth]{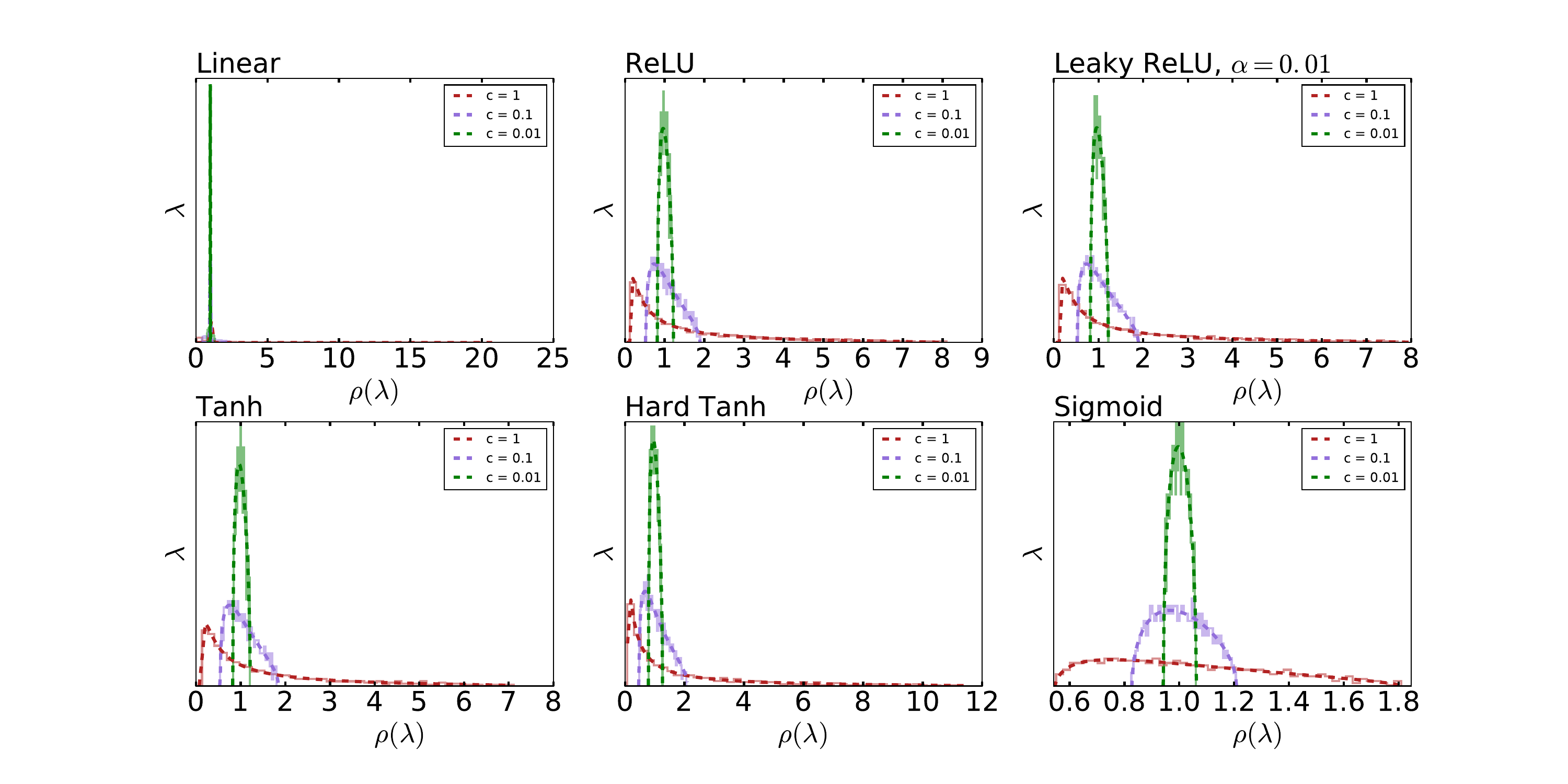}\\
  \caption{Empirical eigenvalue density (solid) and limiting distribution (dashed) from $\J\J^{\T}$ for Gaussian weights with different $\phi(x)$ and $\sigma^2_w = c/L$, $N=400$, $L=100$. }\label{specrum_fig}
\end{figure}

In Fig.~\ref{specrum_fig}, we plot the empirical eigenvalue density (solid) and limiting distribution (dashed) of $\J\J^{\T}$ for Gaussian weights at initialization with  $\sigma^2_w = c/L$. Following activation functions: Linear, ReLU, Leaky ReLU, Tanh, Hard Tanh, and Sigmoid, are explored. The limiting distribution are calculated numerically with  \eqref{stjloc/L} and \eqref{inversion formula}. Note that $\J$ donates the Jacobian matrix of the output of the last residual unit in regard to the input of  first one.

As shown in Fig.~\ref{specrum_fig}, the empirical results agree remarkably  with the theoretical ones. In the cases of $c=1$, different activation functions have diverse  spectral density distributions  because the Jacobian depends  on the signal propagation. Moreover, the smaller $c$, the more concentrated spectrum. In the cases of  the least $c = 0.01$ (green), the  spectral density distributions concentrate around one and their differences   become trivial.

\subsubsection{The training performance}
\begin{figure*}[htbp]
  \centering
  \includegraphics[width=1\textwidth]{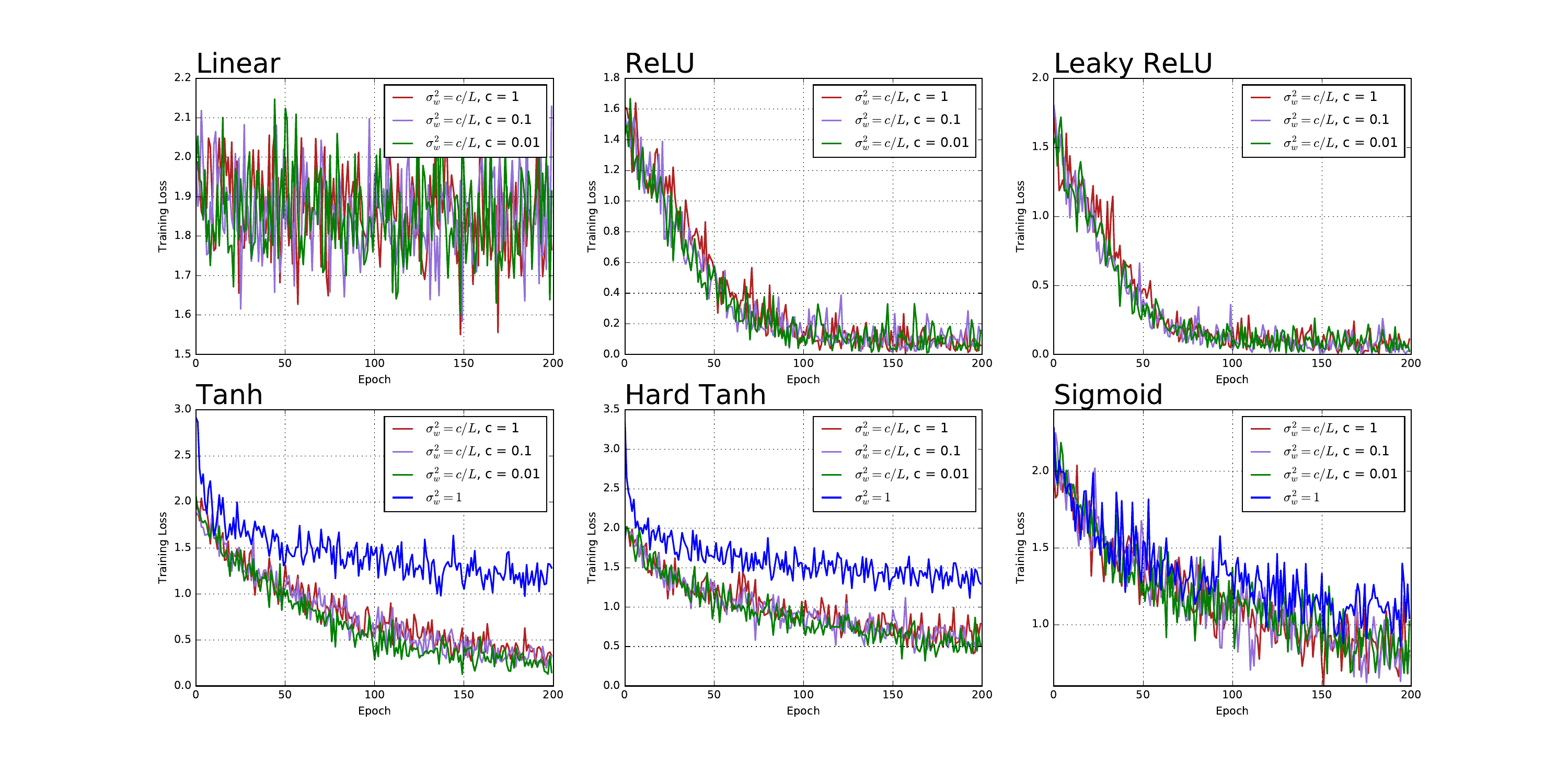}\\
  \caption{Training dynamics of  fully-connected ResNets for different initialization scalings of six common used activation functions with a learning rate of \(10^{-3}\) and momentum $=0.9$.  }\label{fig:Training_dynamics}
\end{figure*}

In Fig.~\ref{fig:Training_dynamics}, we plot the training dynamics   of  fully-connected ResNets for the four initialization scalings  with an optimal learning rate of \(10^{-3}\) and a momentum \(=0.9\). The  training losses of $\sigma_w^2= 1$ are too huge to be included in Fig.~\ref{fig:Training_dynamics}.

As shown in Fig.~\ref{fig:Training_dynamics}, the magnitude of \(\sigma_w^2\)  has a major influence on the learning speed. Based on our theory, an advantage of using layer-dependent scalings of \(\sigma_w^2= c/L\)  is claimed over the classical layer-independent scaling \( \sigma_w^2 = 1 \). This claim is confirmed in Fig.~\ref{fig:Training_dynamics} in which the performances of  \(\sigma_w^2=c/L\)  is much better than those of \(\sigma_w^2=1\) (blue). For the fixed nonlinearity, the smaller $c$, or equivalently,  the more concentrated spectrum becomes at initialization, resulting in the faster learning speed. The optimal training speed is obtained with the least \(c=0.01\) (green), which is the most isometric case. This observation indicates that the stability of the input-output Jacobian spectrum  at initialization strongly predicts the  training  performance, especially at the early stage of training. However, the training speeds vary among experiments with different activation functions   even if  the Jacobian spectrum at initialization are almost the same in the cases of $c=0.01$(see the green line in Fig~\ref{specrum_fig}). This observation indicates that the  Jacobian spectrum at the initialization  is not sufficient to determine the learning performance without  consideration of nonlinearities. Similar results of random scaled orthogonal weights are observed   in Section~\ref{sm-subsec:simu-FC-SGD-Orthogonal} in Appendix.

\subsubsection{The generalization performance} 
\begin{figure*}[htbp]
  \centering
  \includegraphics[width=1\textwidth]{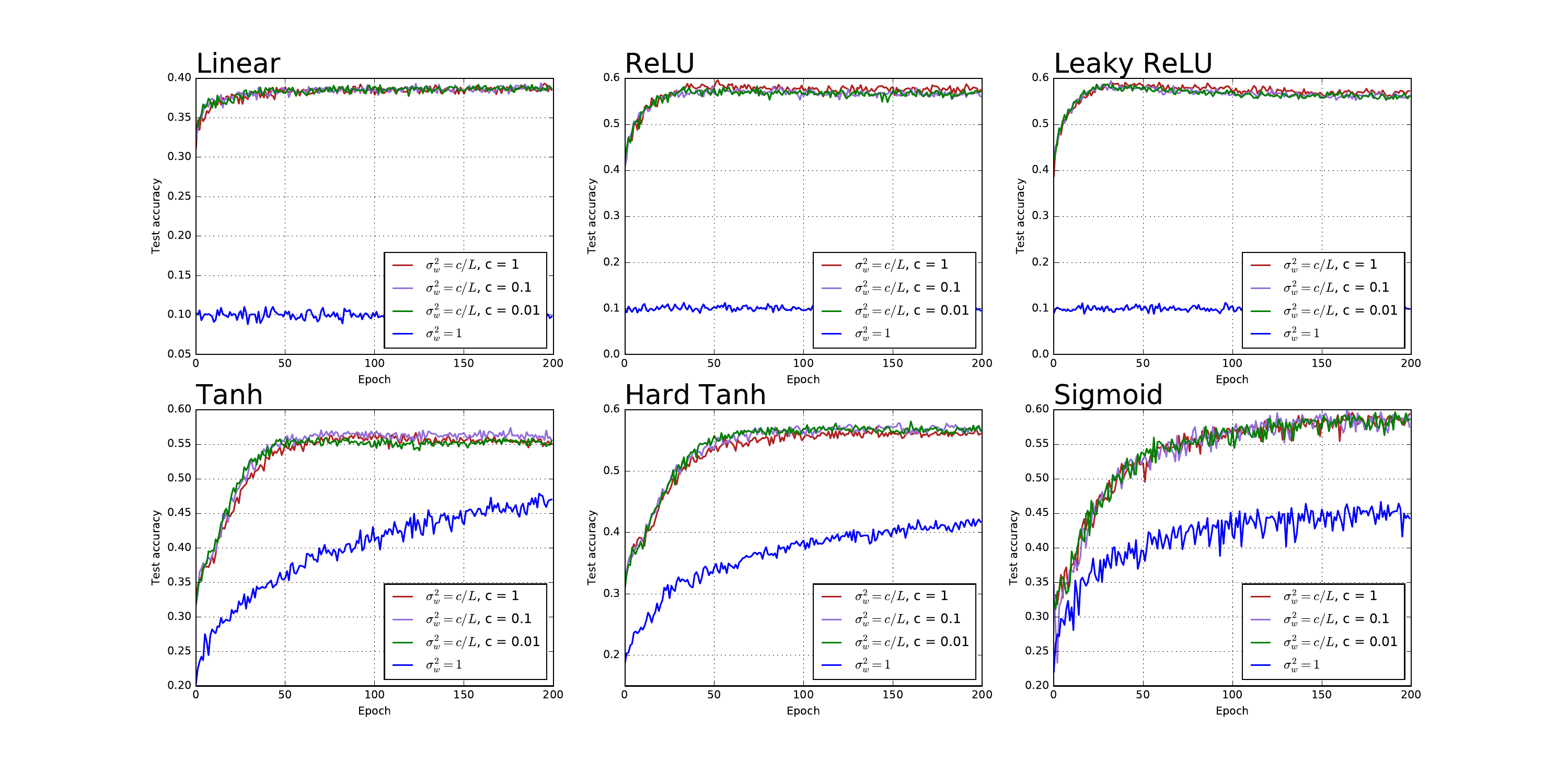}\\
  \caption{The evolutions of test accuracies of  fully-connected ResNets for different initialization scalings of six common used activation functions with a learning rate of \(10^{-3}\) and momentum $=0.9$. }\label{fig:Test_dynamics}
\end{figure*}

We plot the generalization dynamics in Fig.~\ref{fig:Test_dynamics}. Similar with the training dynamics, the generalization performance of $\sigma_w^2 = c/L$ is much better than $\sigma_w^2=1$. For Linear, ReLU and Leaky ReLU, the learning even failures due to the ill-conditioned Jacobian spectrum at the initialization. Moreover, the most concentrated Jacobian spectrum at initialization ($c=0.01$)  is not always the best choice for  generalization performance throughout training. This result indicates that the relationship between  generalization and the Jacobian spectrum goes beyond simply the initialization.




%
\subsection{Convolutional ResNet}
We conduct the experiment based on the convolutional ResNet-110 structure as in~\cite{He2016Identity}. In fact, we can adopt the entire analysis above into the convolutional setting with essentially no modification~\cite{Xiao2018Dynamical}. In this convolutional ResNet, each residual unit contains a shallow network of \(2\) layers. Therefore, by considering the Jacobian of \textit{all} residual units  three scalings of \(\sigma_w^2= 1,1/\sqrt{L}\) and \(0.01/\sqrt{L}\) are tested. In this experiment, we focus on the learning speed at the early stage of training. For the three choices of \(\sigma_w^2\), the best performance is always achieved by the learning rate of \(10^{-2}\) with momentum \(=0.9\)  and is presented in Fig.~\ref{fig:gauss-orth-training-speed-conv}. We refer the reader to Section~\ref{sm-subsec:simu-conv} in Appendix for the results of the ADAM optimizer. We observe from Fig.~\ref{fig:gauss-orth-training-speed-conv} that: 1) for convolutional ResNet without BN, the magnitude of \(\sigma_w^2\) plays a central role in obtaining a satisfying learning speed. The optimal learning speed is achieved with \(\sigma^2=1/\sqrt{L}\), while large \(\sigma^2=1\) without BN results in exploding gradient and hence the failure of training.  Surprisingly, small \(\sigma^2=1/\sqrt{L}\) without the regularization effect of the BN \cite{ioffe2015batch} \textit{still} has roughly the same performance as \(\sigma^2=1\) with BN.
2) it is noteworthy that, different from the fully-connected ResNet case, for the convolutional ResNet of \(2\) layers in each residual unit, the scaling \(\sigma^2=1/\sqrt{L}\) always outperforms \(\sigma^2=0.01/\sqrt{L}\) while the latter achieves the \textit{dynamical isometry}. This observation is not trivial and comes possibly from the fact that extremely small \(\sigma_w^2\) causes the internal gradient vanishing inside each residual unit. We will discuss this point in the next section.

%
%
%
%
%
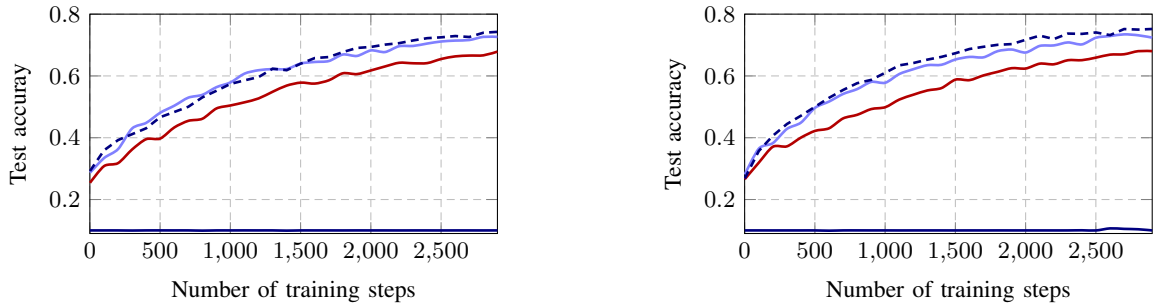
\begin{figure}[htb]
	\centering
	\begin{subfigure}[t]{0.48\textwidth}
		\centering
		\begin{tikzpicture}[font=\small,spy using outlines]
		\renewcommand{\axisdefaulttryminticks}{4}
		\pgfplotsset{every major grid/.style={densely dashed}}
		\tikzstyle{every axis y label}+=[yshift=-10pt]
		\tikzstyle{every axis x label}+=[yshift=5pt]
		\begin{axis}[
		width=7cm,
		height=4.5cm,
		xmin=0,
		ymin=0.09,
		xmax=2901,
		ymax=0.8,
		grid=major,
		ymajorgrids=true,
		scaled ticks=true,
		xlabel={Number of training steps},
		ylabel={Test accuray}
		]
		\addplot[smooth,mycolor1,line width=1pt] plot coordinates{
			(1.000000,0.254000)(101.000000,0.308133)(201.000000,0.317867)(301.000000,0.362100)(401.000000,0.395333)(501.000000,0.397400)(601.000000,0.433267)(701.000000,0.454700)(801.000000,0.461433)(901.000000,0.494733)(1001.000000,0.504433)(1101.000000,0.514633)(1201.000000,0.527167)(1301.000000,0.548767)(1401.000000,0.568733)(1501.000000,0.578300)(1601.000000,0.575200)(1701.000000,0.585900)(1801.000000,0.608600)(1901.000000,0.605933)(2001.000000,0.617867)(2101.000000,0.630700)(2201.000000,0.642600)(2301.000000,0.641000)(2401.000000,0.641467)(2501.000000,0.654900)(2601.000000,0.663300)(2701.000000,0.665900)(2801.000000,0.666633)(2901.000000,0.678700)
		};
		\addplot[smooth,mycolor2,line width=1pt] plot coordinates{
			(1.000000,0.286667)(101.000000,0.334200)(201.000000,0.363267)(301.000000,0.430067)(401.000000,0.448133)(501.000000,0.480933)(601.000000,0.503600)(701.000000,0.529767)(801.000000,0.538833)(901.000000,0.563633)(1001.000000,0.579033)(1101.000000,0.607533)(1201.000000,0.618500)(1301.000000,0.622100)(1401.000000,0.622300)(1501.000000,0.640200)(1601.000000,0.645233)(1701.000000,0.648167)(1801.000000,0.669367)(1901.000000,0.664700)(2001.000000,0.682633)(2101.000000,0.677300)(2201.000000,0.696200)(2301.000000,0.697067)(2401.000000,0.704800)(2501.000000,0.711533)(2601.000000,0.714400)(2701.000000,0.716267)(2801.000000,0.726433)(2901.000000,0.726367)
		};
		\addplot[smooth,mycolor3,line width=1pt] plot coordinates{
			(1.000000,0.100000)(101.000000,0.100000)(201.000000,0.100000)(301.000000,0.099616)(401.000000,0.100043)(501.000000,0.100093)(601.000000,0.100023)(701.000000,0.100049)(801.000000,0.099321)(901.000000,0.100003)(1001.000000,0.100000)(1101.000000,0.100093)(1201.000000,0.100023)(1301.000000,0.100049)(1401.000000,0.099321)(1501.000000,0.100003)(1601.000000,0.100000)(1701.000000,0.100000)(1801.000000,0.099933)(1901.000000,0.100033)(2001.000000,0.100000)(2101.000000,0.100000)(2201.000000,0.099933)(2301.000000,0.100000)(2401.000000,0.100000)(2501.000000,0.099933)(2601.000000,0.100000)(2701.000000,0.100000)(2801.000000,0.099933)(2901.000000,0.100000)
		};
		\addplot[densely dashed,mycolor3,line width=1pt] plot coordinates{
			(1.000000,0.292333)(101.000000,0.359267)(201.000000,0.392467)(301.000000,0.411500)(401.000000,0.430633)(501.000000,0.466600)(601.000000,0.483333)(701.000000,0.500333)(801.000000,0.530633)(901.000000,0.551967)(1001.000000,0.574167)(1101.000000,0.585700)(1201.000000,0.596333)(1301.000000,0.623433)(1401.000000,0.618567)(1501.000000,0.639800)(1601.000000,0.657867)(1701.000000,0.661133)(1801.000000,0.677300)(1901.000000,0.690533)(2001.000000,0.693900)(2101.000000,0.701367)(2201.000000,0.706033)(2301.000000,0.714900)(2401.000000,0.722200)(2501.000000,0.725000)(2601.000000,0.729100)(2701.000000,0.726033)(2801.000000,0.739833)(2901.000000,0.742667)
		};
		\end{axis}
		\end{tikzpicture}
		\caption{Gaussian weights with \(\W_{ij}^l \sim \mathcal{N}(0,\sigma_w^2/N) \) }
		\label{subfig:gauss-training-speed-conv}
	\end{subfigure}%
	\begin{subfigure}[t]{0.48\textwidth}
		\centering
		\begin{tikzpicture}[font=\small,spy using outlines]
		\renewcommand{\axisdefaulttryminticks}{4}
		\pgfplotsset{every major grid/.style={densely dashed}}
		\tikzstyle{every axis y label}+=[yshift=-10pt]
		\tikzstyle{every axis x label}+=[yshift=5pt]
		\begin{axis}[
		width=7cm,
		height=4.5cm,
		xmin=0,
		ymin=0.09,
		xmax=2901,
		ymax=0.8,
		grid=major,
		ymajorgrids=true,
		scaled ticks=true,
		xlabel={Number of training steps},
		ylabel={Test accuracy}
		]
		\addplot[smooth,mycolor1,line width=1pt] plot coordinates{
			(1.000000,0.265767)(101.000000,0.319433)(201.000000,0.370733)(301.000000,0.372067)(401.000000,0.400500)(501.000000,0.422167)(601.000000,0.430533)(701.000000,0.462567)(801.000000,0.474167)(901.000000,0.492733)(1001.000000,0.498800)(1101.000000,0.523533)(1201.000000,0.539267)(1301.000000,0.553033)(1401.000000,0.560667)(1501.000000,0.588100)(1601.000000,0.586433)(1701.000000,0.601867)(1801.000000,0.613233)(1901.000000,0.625100)(2001.000000,0.624000)(2101.000000,0.639767)(2201.000000,0.637700)(2301.000000,0.650900)(2401.000000,0.651033)(2501.000000,0.659333)(2601.000000,0.668933)(2701.000000,0.670167)(2801.000000,0.680200)(2901.000000,0.680300)
		};
		\addplot[smooth,mycolor2,line width=1pt] plot coordinates{
			(1.000000,0.280933)(101.000000,0.364700)(201.000000,0.382600)(301.000000,0.428467)(401.000000,0.449000)(501.000000,0.496800)(601.000000,0.517067)(701.000000,0.541700)(801.000000,0.557367)(901.000000,0.580700)(1001.000000,0.577967)(1101.000000,0.606233)(1201.000000,0.621833)(1301.000000,0.634800)(1401.000000,0.636100)(1501.000000,0.653333)(1601.000000,0.661533)(1701.000000,0.660333)(1801.000000,0.680233)(1901.000000,0.685433)(2001.000000,0.675833)(2101.000000,0.697267)(2201.000000,0.698767)(2301.000000,0.708433)(2401.000000,0.702100)(2501.000000,0.723667)(2601.000000,0.729167)(2701.000000,0.734567)(2801.000000,0.731567)(2901.000000,0.724033)
		};
		\addplot[smooth,mycolor3,line width=1pt] plot coordinates{
			(1.000000,0.100013)(101.000000,0.100025)(201.000000,0.099963)(301.000000,0.100000)(401.000000,0.100000)(501.000000,0.100000)(601.000000,0.099157)(701.000000,0.100043)(801.000000,0.100093)(901.000000,0.100023)(1001.000000,0.100049)(1101.000000,0.099932)(1201.000000,0.100003)(1301.000000,0.100000)(1401.000000,0.100000)(1501.000000,0.099933)(1601.000000,0.100000)(1701.000000,0.099967)(1801.000000,0.100033)(1901.000000,0.100000)(2001.000000,0.100000)(2101.000000,0.100000)(2201.000000,0.099967)(2301.000000,0.100000)(2401.000000,0.100433)(2501.000000,0.099901)(2601.000000,0.106567)(2701.000000,0.104967)(2801.000000,0.103733)(2901.000000,0.100000)
		};
		\addplot[densely dashed,mycolor3,line width=1pt] plot coordinates{
			(1.000000,0.268250)(101.000000,0.357100)(201.000000,0.406500)(301.000000,0.444350)(401.000000,0.471250)(501.000000,0.500200)(601.000000,0.529250)(701.000000,0.554600)(801.000000,0.576500)(901.000000,0.587250)(1001.000000,0.610450)(1101.000000,0.634300)(1201.000000,0.641800)(1301.000000,0.652800)(1401.000000,0.661700)(1501.000000,0.673600)(1601.000000,0.687250)(1701.000000,0.694350)(1801.000000,0.700750)(1901.000000,0.703050)(2001.000000,0.717050)(2101.000000,0.730450)(2201.000000,0.719400)(2301.000000,0.737100)(2401.000000,0.736150)(2501.000000,0.740700)(2601.000000,0.732150)(2701.000000,0.751750)(2801.000000,0.750000)(2901.000000,0.751950)
		};
		\end{axis}
		\end{tikzpicture}
		\caption{Orthogonal weights with \( \W^l (\W^l)^\T=\sigma_w^2\I_N\) }
		\label{subfig:orth-training-speed-conv}
	\end{subfigure}%
	\caption{ { Learning dynamics of a convolutional ResNet for different initialization scalings \( \sigma_w^2 = 0.01/\sqrt{L} \) (red), \( \sigma_w^2 = 1/\sqrt{L} \) (purple) and \( \sigma_w^2 = 1 \) (dark blue), with a learning rate of \(10^{-2}\) and the ReLU nonlinearity. Solid lines without BN and dashed ones with the BN procedure added (for \( \sigma_w^2 = 1 \)). } }
	\label{fig:gauss-orth-training-speed-conv}
\end{figure}

\section{Discussion and Conclusion}
\label{sec:dis-conclu}
In this article, exploiting advanced tools in free probability in the regime of a large network, we establish, that for ResNets, the variance of the initial random weights should {\it also} be scaled as a function of the number of layers.
In particular, the theoretical results show that for large \(L\) the condition for spectrum concentration is in fact \textit{universal} in the sense that, for almost \emph{all} of the common-used nonlinearities    and both weight initialization methods (Gaussian or orthogonal), it is sufficient and necessary to take \( \sigma_w^2 = O \left( 1/L \right) \) to ensure the squared singular values of the input-output Jacobian to be of order $O(1)$ (thus is neither vanishing nor exploding). The weights scaling essentially results in the \textit{eigenspectrum concentration} of \(\J\J^\T\), such that the error vector will be properly preserved under backpropagation.
We then provide in Section~\ref{sec:experiments} the comparison of empirical evidences with our theoretical results. Mathematically speaking, our approach holds only asymptotically as \(N,L \to \infty\), practical advantages are observed for finite width  \(N\) and depth \(L\), when applied to the popular CIFAR-10 dataset, for both fully-connected and convolutional ResNets. This agreement is not surprising, as observed in many other fields~\cite{Silverstein2010Spectral,plerou1999universal,liao2018on}.

In  practice, a residual unit  always contains a shallow network with \(m\) layers. In this way, the dynamic of the ResNet is given by
\begin{equation}
  \begin{cases}
    \h_1^l &=  \W_1^l\x^{l-1} + \bb_1^l,\\
    \h_2^l &=  \W_2^l\phi(\h_{1}^l) + \bb_2^l,\\
     &\cdots\\
    \h_m^l &=  \W_{m}^l\phi(\h_{m-1}^l) + \bb_m^l,\\
        \x^l &= \x^{l-1} + \phi(\h_m^l).
  \end{cases}
\end{equation}
The entire input-output Jacobian matrix is \(\J= \prod\limits_{l=1}^L {(\I_N+\hat{\J}_l)}\), where \(\hat{\J}_l= \frac{\partial \phi(\h_m^L)}{\partial \x^{l-1}}= \prod\limits_{i = 1}^m {\D_i^l\W_i^l}\) denotes the local Jacobian matrix of a residual unit. For \(m>1\), let \(\sigma^2=O(L^{-1/m})\) and the spectrum of  \(\J\J^\T\) will be well-conditioned. However, the eigenvalues of  the local Jacobian matrix \(\hat{\J}_l\hat{\J}_l^\T\) will  be extremely  small and  cause the gradient vanishing in the local residual unit. Thus,  a \textit{trade-off}  between the entire and local input-output Jacobian matrix is always required.

In future work, it would be interesting to extend our theoretical framework to more general skip connections. Moreover, exploring new weight initializations or nonlinearities to handle the \textit{trade-off} mentioned above would be of practical significance.

\bibliographystyle{IEEEtran}
\bibliography{RMT4ResNet}

\cleardoublepage
\appendix
\subsection{Proof of Theorem.~\ref{subsec:multi-layer-case}}
\begin{Proof}[Multi-layer case]
\label{proof:Multi-layer}
\normalfont Since
\begin{equation}
M_{\J_l\J^\T_l}=\frac{m^{(l)}_1}{z}+\frac{m^{(l)}_2}{z^2}+\cdots.
\end{equation}
Employ the Lagrange inversion theorem~\cite{Ferraro2008Lagrange} and we have,
\begin{equation}
M^{-1}_{\J_l\J^\T_l}=\frac{m^{(l)}_1}{z}+\frac{m^{(l)}_2}{m^{(l)}_1}+\cdots.
\end{equation}
According to \eqref{s*s},
\begin{equation}
S_{\J_l\J^\T_l}=\frac{1+z}{zM^{-1}_{\J_l\J^\T_l}}=\frac{1}{m^{(l)}_1}+\left(\frac{1}{m^{(l)}_1}-\frac{m^{(l)}_2}{(m^{(l)}_1)^3}\right)z+\cdots,
\end{equation}
Thus,
\begin{equation}
\begin{split}
&S_{\J\J^\T}= \prod_{l=1}^{L}S_{\J_l\J^\T_l}\\
&=\prod_{l=1}^{L}\frac{1}{m^{(l)}_1}+ \sum_{j=1}^{L} \left[\left(\frac{\left(m^{(j)}_1\right)^2-m^{(j)}_2}{\left(m^{(j)}_1\right)^2}\right)\prod_{l= 1}^{L}\frac{1}{m^{(l)}_1}\right]z+\cdots.
\end{split}
\end{equation}
For the sake of simplicity, let \(X\equiv \prod_{l=1}^{L}\frac{1}{m^{(l)}_1}\) and \(Y\equiv \sum_{j=1}^{L} \left[\left(\frac{\left(m^{(j)}_1\right)^2-m^{(j)}_2}{\left(m^{(j)}_1\right)^2}\right)\prod_{l=1}\frac{1}{m^{(l)}_1}\right]\). Expand \(M^{-1}_{\J\J^\T}\) and we have,
\begin{equation}
M^{-1}_{\J\J^\T}=\frac{1+z}{zS_{\J\J^\T}}=\frac{1}{X z} + \frac{X - Y}{X^2}+\cdots.
\end{equation}
Let \(m^k:=\int {d\lambda \rho_{\J\J^\T}(\lambda)\lambda^k}\), we obtain the following equations:
\begin{equation}
m_1 =\frac{1}{X}=\prod_{l=1}^{L}{m^{(l)}_1},\quad m_2 =\frac{X - Y}{X^3}.
\end{equation}
Finally, we obtain the expectation \(\mu_{\J\J^\T}\) and variance \(\sigma^2_{\J\J^\T}\) of \(\J\J^\T\),
\begin{equation}
\begin{split}
\mu_{\J\J^\T}&=m_1=\prod_{l=1}^{L}{m^{(l)}_1},\\
\sigma^2_{\J\J^\T}&=m_2-m_1^2=\left(\prod_{l=1}^{L}{m^{(l)}_1}\right)^2\sum_{l=1}^{L} \frac{m^{(l)}_2-\left(m^{(l)}_1\right)^2}{\left(m^{(l)}_1\right)^2}.
\end{split}
\label{mu_temp}
\end{equation}
\end{Proof}

\subsection{Fully-connected ResNets: SGD-Momentum with Random Scaled Orthogonal Weights}
\label{sm-subsec:simu-FC-SGD-Orthogonal}
\begin{figure}[ht]
	\centering
	\includegraphics[width=1\textwidth]{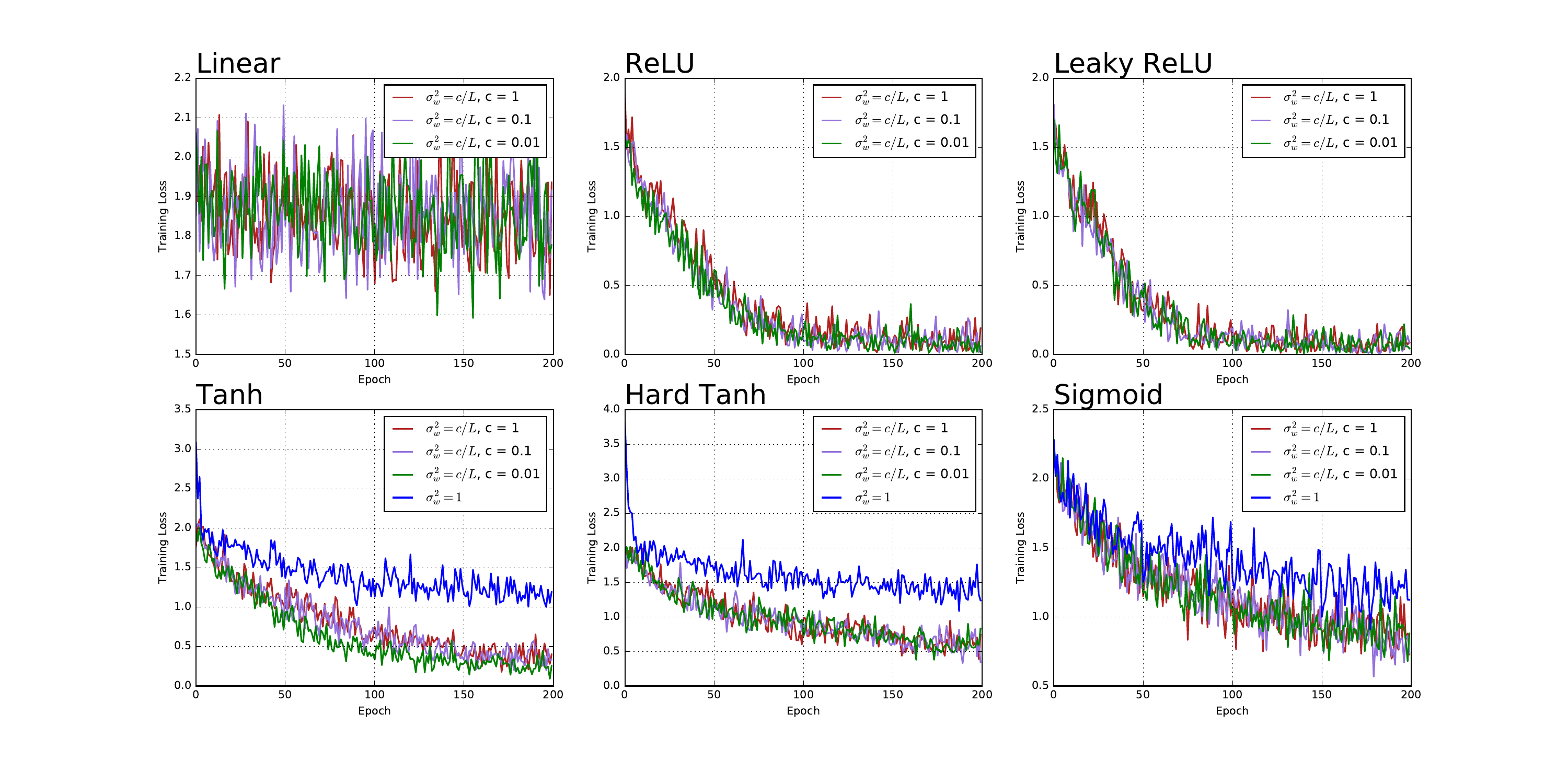}\\
	\caption{Training dynamics of  fully-connected ResNets for the different initialization scalings of scaled random orthogonal weights. Six common used activation functions are tested. The optimizer is SGD-Momentum with an  learning rate of \(10^{-3}\) and a momentum \(=0.9\).}\label{fig:Training_dynamics_SGD-Orthogonal}
\end{figure}

\begin{figure}[ht]
	\centering
	\includegraphics[width=1\textwidth]{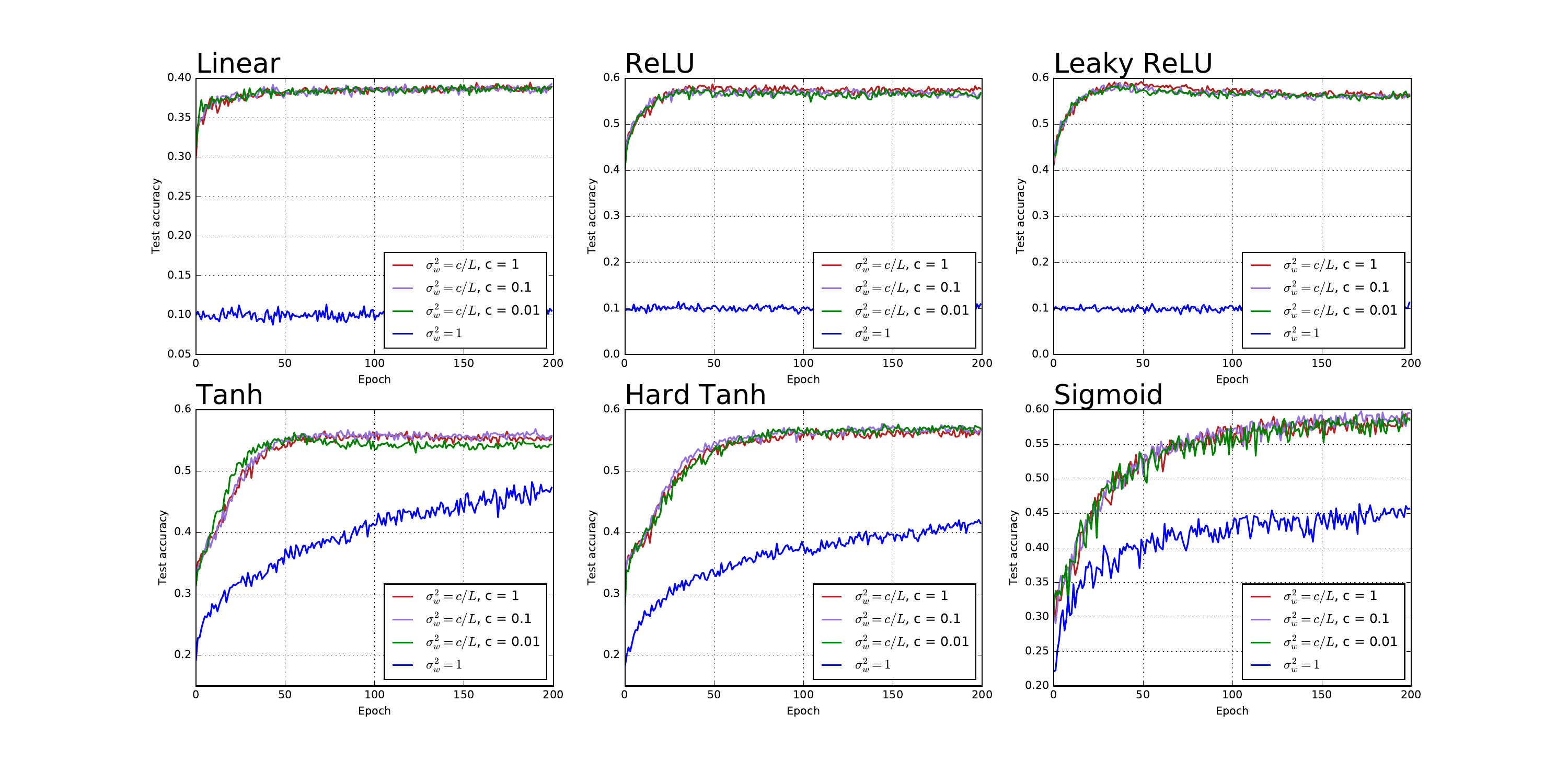}\\
	\caption{The evolutions of test accuracies of  fully-connected ResNets for the different initialization scalings of scaled random orthogonal weights. Six common used activation functions are tested. The optimizer is SGD-Momentum with an  learning rate of \(10^{-3}\) and a momentum \(=0.9\).}\label{fig:Test_dynamics_SGD-Orthogonal}
\end{figure}

In Fig.~\ref{fig:Training_dynamics_SGD-Orthogonal}, we plot the training dynamics  of  fully-connected ResNets  for the four initialization scalings of scaled random orthogonal weights with six common used activation functions. The optimizer is SGD-Momentum with an optimal learning rate of \(10^{-3}\) and a momentum \(=0.9\). In Fig.~\ref{fig:Test_dynamics_SGD-Orthogonal}, we plot the evolutions of test accuracies.
\subsection{Fully-connected ResNets: ADAM }
\label{sm-sec:simu}
\subsubsection{Random Scaled Gaussian Weights}
\begin{figure}[ht]
	\centering
	\includegraphics[width=1\textwidth]{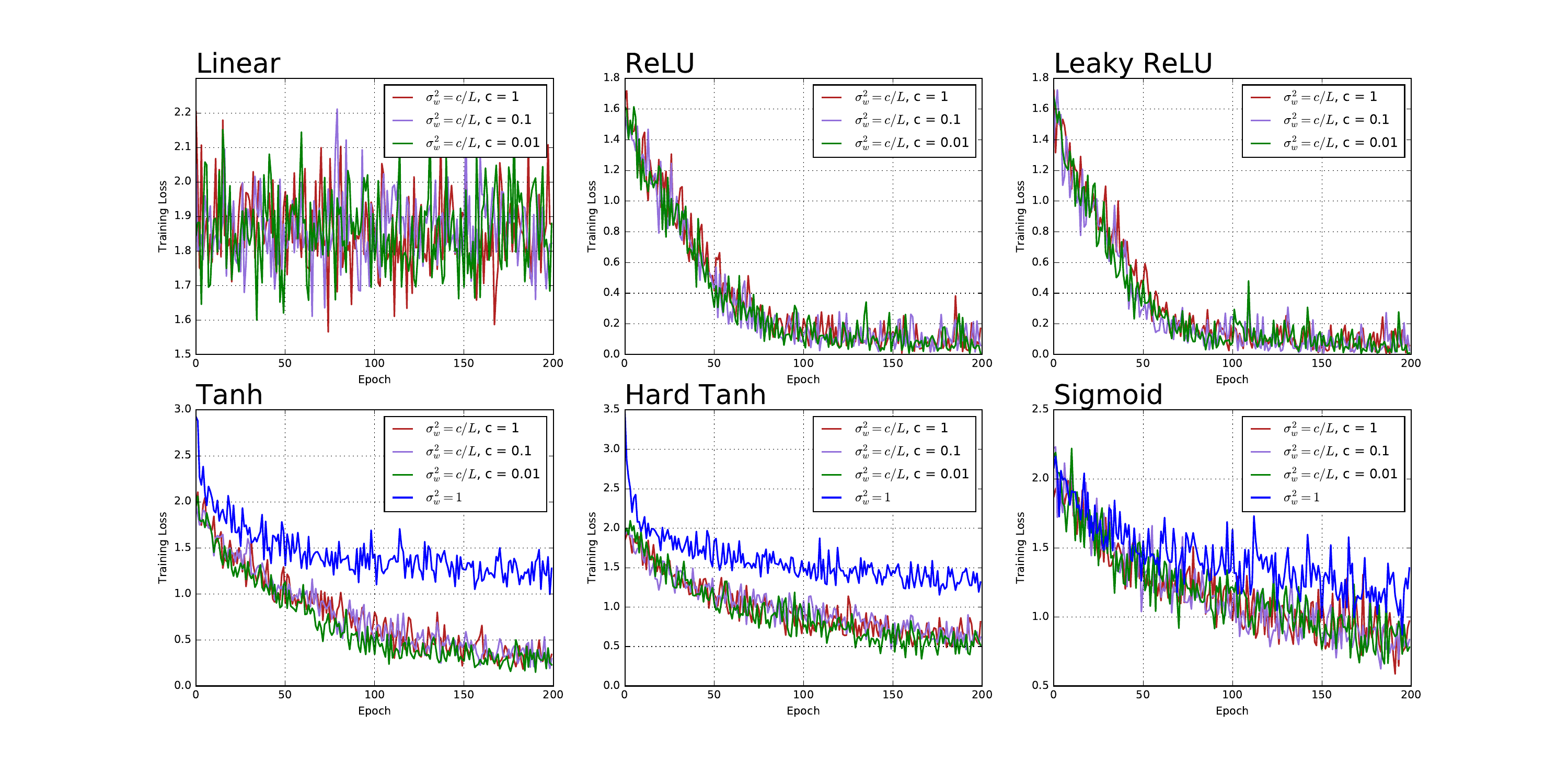}\\
	\caption{Training dynamics of  fully-connected ResNets for the different initialization scalings of scaled random Gaussian weights. Six common used activation functions are tested. The optimizer is ADAM with an  initial  rate of \(10^{-4}\).}\label{fig:Training_dynamics_ADAM}
\end{figure}

\begin{figure}[ht]
	\centering
	\includegraphics[width=1\textwidth]{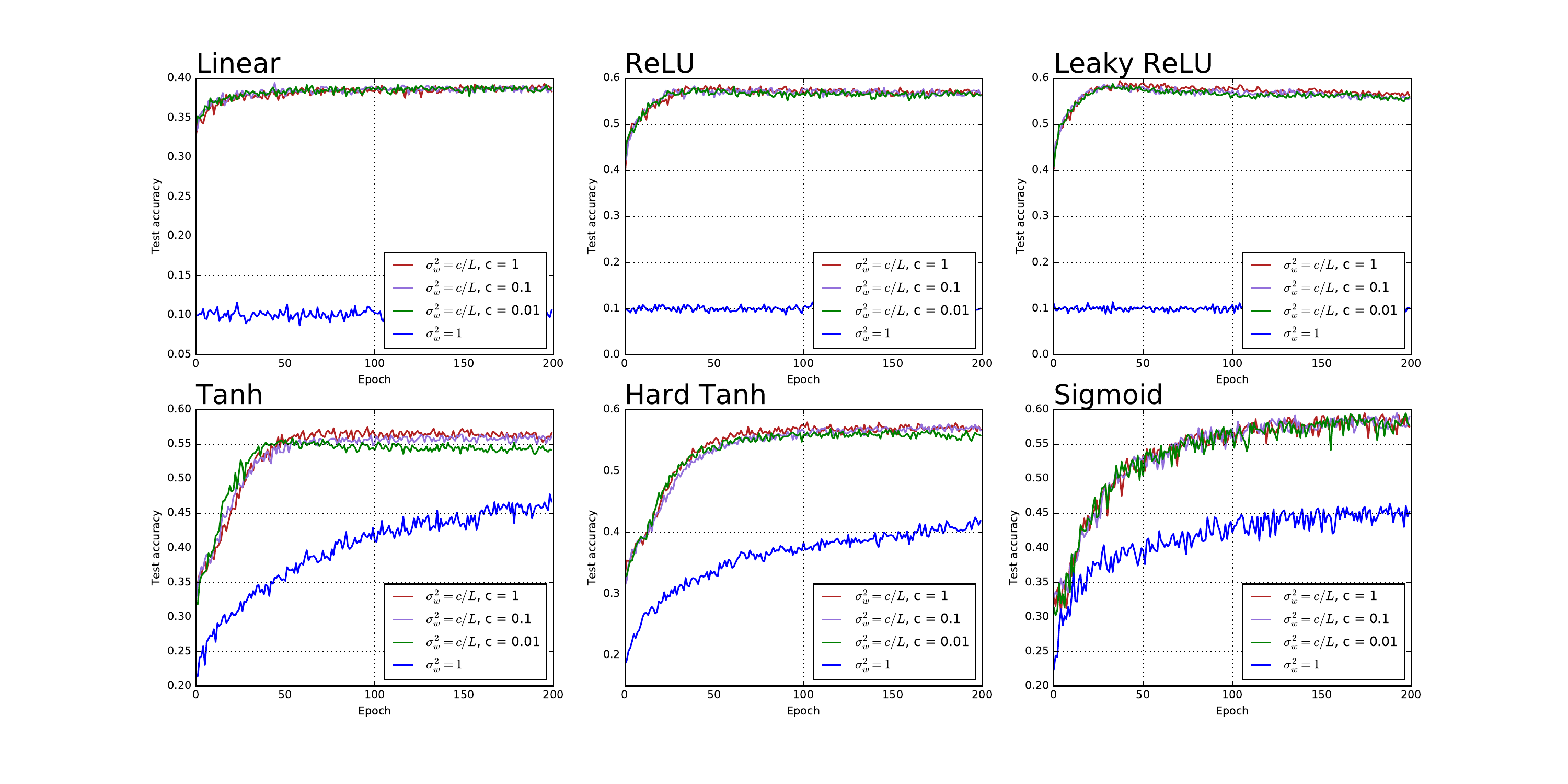}\\
	\caption{The evolutions of test accuracies fully-connected ResNets for the different initialization scalings of scaled random Gaussian weights. Six common used activation functions are tested. The optimizer is ADAM with an  initial  rate of \(10^{-4}\). }\label{fig:Test_dynamics_ADAM}
\end{figure}

In Fig.~\ref{fig:Training_dynamics_ADAM}, we plot the training dynamics  of  fully-connected ResNets  for the four initialization scalings of scaled random Gaussian weights with six common used activation functions. The optimizer is ADAM with an initial learning rate of \(10^{-4}\). In Fig.~\ref{fig:Test_dynamics_ADAM}, we plot the evolutions of test accuracies.

\subsubsection{Random Scaled Orthogonal Weights}
\begin{figure}[htb]
	\centering
	\includegraphics[width=1\textwidth]{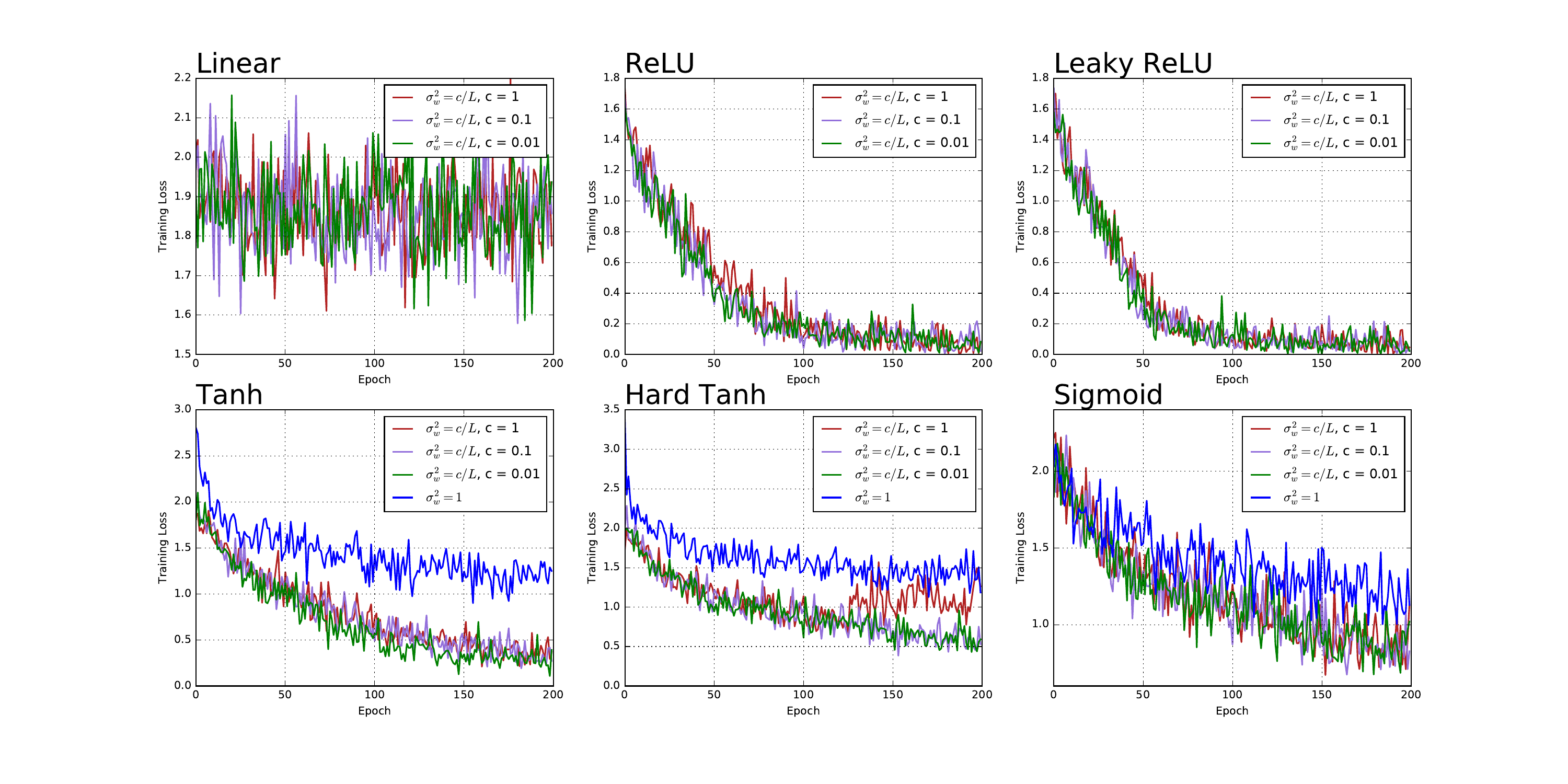}\\
	\caption{Training dynamics of  fully-connected ResNets for the different initialization scalings of scaled random orthogonal weights. Six common used activation functions are tested. The optimizer is ADAM with an  initial  rate of \(10^{-4}\).}\label{fig:Training_dynamics_ADAM_orth}
\end{figure}

\begin{figure}[htb]
	\centering
	\includegraphics[width=1\textwidth]{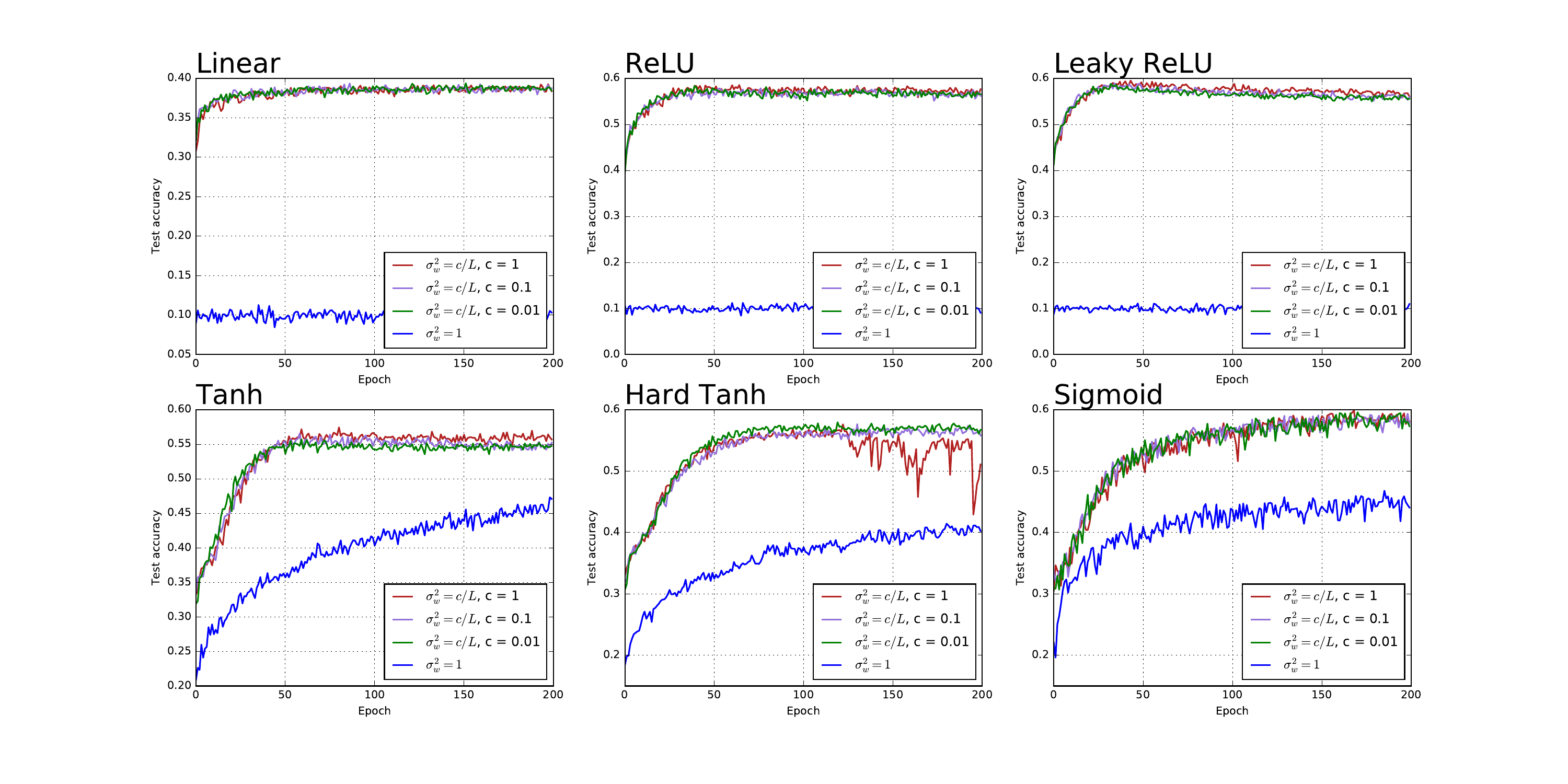}\\
	\caption{The evolutions of test accuracies  of  fully-connected ResNets for the different initialization scalings of scaled random orthogonal weights. Six common used activation functions are tested. The optimizer is ADAM with an  initial  rate of \(10^{-4}\).}\label{fig:Test_dynamics_ADAM_orth}
\end{figure}

In Fig.~\ref{fig:Training_dynamics_ADAM_orth}, we plot the training dynamics  of  fully-connected ResNets  for the four initialization scalings of scaled random orthogonal weights with six common used activation functions. The optimizer is ADAM with an initial learning rate of \(10^{-4}\). In Fig.~\ref{fig:Test_dynamics_ADAM_orth}, we plot the evolutions of test accuracies.
\subsection{Convolutional ResNet}
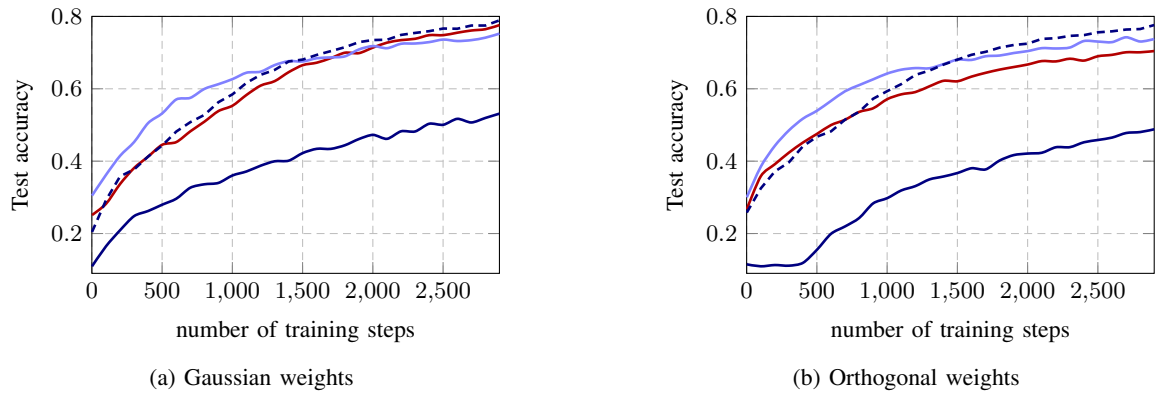
\begin{figure}[ht]
	\centering
	\begin{subfigure}[t]{0.48\textwidth}
		\centering
		\begin{tikzpicture}[font=\small,spy using outlines]
		\renewcommand{\axisdefaulttryminticks}{4}
		\pgfplotsset{every major grid/.style={densely dashed}}
		\tikzstyle{every axis y label}+=[yshift=-10pt]
		\tikzstyle{every axis x label}+=[yshift=5pt]
		\begin{axis}[
		width=7cm,
		height=5cm,
		xmin=0,
		ymin=0.09,
		xmax=2901,
		ymax=0.8,
		grid=major,
		ymajorgrids=true,
		scaled ticks=true,
		xlabel={number of training steps},
		ylabel={Test accuracy}
		]
		\addplot[smooth,mycolor1,line width=1pt] plot coordinates{
			(1.000000,0.251133)(101.000000,0.282167)(201.000000,0.336967)(301.000000,0.381233)(401.000000,0.412733)(501.000000,0.445400)(601.000000,0.452767)(701.000000,0.482467)(801.000000,0.509100)(901.000000,0.538467)(1001.000000,0.553267)(1101.000000,0.583133)(1201.000000,0.608733)(1301.000000,0.621100)(1401.000000,0.645967)(1501.000000,0.665300)(1601.000000,0.672133)(1701.000000,0.684767)(1801.000000,0.699667)(1901.000000,0.698533)(2001.000000,0.713933)(2101.000000,0.727533)(2201.000000,0.734500)(2301.000000,0.738133)(2401.000000,0.748167)(2501.000000,0.748167)(2601.000000,0.754967)(2701.000000,0.761067)(2801.000000,0.764600)(2901.000000,0.776000)
		};
		\addplot[smooth,mycolor2,line width=1pt] plot coordinates{
			(1.000000,0.304733)(101.000000,0.362300)(201.000000,0.414800)(301.000000,0.452967)(401.000000,0.505467)(501.000000,0.532067)(601.000000,0.570533)(701.000000,0.575000)(801.000000,0.599700)(901.000000,0.612767)(1001.000000,0.626733)(1101.000000,0.644500)(1201.000000,0.647300)(1301.000000,0.665500)(1401.000000,0.676367)(1501.000000,0.675467)(1601.000000,0.684467)(1701.000000,0.686667)(1801.000000,0.689600)(1901.000000,0.708733)(2001.000000,0.718033)(2101.000000,0.712133)(2201.000000,0.724433)(2301.000000,0.725133)(2401.000000,0.728967)(2501.000000,0.736000)(2601.000000,0.731933)(2701.000000,0.734467)(2801.000000,0.741233)(2901.000000,0.752467)
		};
		\addplot[smooth,mycolor3,line width=1pt] plot coordinates{
			(1.000000,0.109333)(101.000000,0.164967)(201.000000,0.209167)(301.000000,0.248267)(401.000000,0.262367)(501.000000,0.279833)(601.000000,0.296300)(701.000000,0.326633)(801.000000,0.336133)(901.000000,0.339900)(1001.000000,0.360533)(1101.000000,0.371600)(1201.000000,0.387467)(1301.000000,0.399667)(1401.000000,0.401367)(1501.000000,0.422467)(1601.000000,0.434133)(1701.000000,0.434000)(1801.000000,0.443967)(1901.000000,0.461167)(2001.000000,0.472633)(2101.000000,0.461733)(2201.000000,0.482533)(2301.000000,0.482100)(2401.000000,0.503400)(2501.000000,0.500567)(2601.000000,0.516967)(2701.000000,0.506933)(2801.000000,0.519400)(2901.000000,0.531300)
		};
		\addplot[densely dashed,mycolor3,line width=1pt] plot coordinates{
			(1.000000,0.204267)(101.000000,0.292533)(201.000000,0.358033)(301.000000,0.377233)(401.000000,0.413067)(501.000000,0.443467)(601.000000,0.481867)(701.000000,0.507800)(801.000000,0.527867)(901.000000,0.563000)(1001.000000,0.584733)(1101.000000,0.616467)(1201.000000,0.638500)(1301.000000,0.652100)(1401.000000,0.675467)(1501.000000,0.680767)(1601.000000,0.693667)(1701.000000,0.704100)(1801.000000,0.714867)(1901.000000,0.728933)(2001.000000,0.734733)(2101.000000,0.735533)(2201.000000,0.748900)(2301.000000,0.754400)(2401.000000,0.759200)(2501.000000,0.766500)(2601.000000,0.765933)(2701.000000,0.774800)(2801.000000,0.774800)(2901.000000,0.789100)
		};
		\end{axis}
		\end{tikzpicture}
		\caption{Gaussian weights }
		\label{subfig:gauss-training-speed-conv-adam}
	\end{subfigure}%
	\begin{subfigure}[t]{0.48\textwidth}
		\centering
		\begin{tikzpicture}[font=\small,spy using outlines]
		\renewcommand{\axisdefaulttryminticks}{4}
		\pgfplotsset{every major grid/.style={densely dashed}}
		\tikzstyle{every axis y label}+=[yshift=-10pt]
		\tikzstyle{every axis x label}+=[yshift=5pt]
		\begin{axis}[
		width=7cm,
		height=5cm,
		xmin=0,
		ymin=0.09,
		xmax=2901,
		ymax=0.8,
		grid=major,
		ymajorgrids=true,
		scaled ticks=true,
		xlabel={number of training steps},
		ylabel={Test accuracy}
		]
		\addplot[smooth,mycolor1,line width=1pt] plot coordinates{
			(1.000000,0.267000)(101.000000,0.358500)(201.000000,0.391600)(301.000000,0.423633)(401.000000,0.451500)(501.000000,0.475433)(601.000000,0.499867)(701.000000,0.514533)(801.000000,0.536100)(901.000000,0.546300)(1001.000000,0.571400)(1101.000000,0.584700)(1201.000000,0.590967)(1301.000000,0.606833)(1401.000000,0.622067)(1501.000000,0.620800)(1601.000000,0.633700)(1701.000000,0.644133)(1801.000000,0.653033)(1901.000000,0.660100)(2001.000000,0.667133)(2101.000000,0.676600)(2201.000000,0.675667)(2301.000000,0.683067)(2401.000000,0.677933)(2501.000000,0.690000)(2601.000000,0.693833)(2701.000000,0.700933)(2801.000000,0.700867)(2901.000000,0.704267)
		};
		\addplot[smooth,mycolor2,line width=1pt] plot coordinates{
			(1.000000,0.299900)(101.000000,0.384933)(201.000000,0.442533)(301.000000,0.485067)(401.000000,0.517567)(501.000000,0.539433)(601.000000,0.566633)(701.000000,0.593300)(801.000000,0.610167)(901.000000,0.626200)(1001.000000,0.642233)(1101.000000,0.652800)(1201.000000,0.657200)(1301.000000,0.656667)(1401.000000,0.668100)(1501.000000,0.680600)(1601.000000,0.679767)(1701.000000,0.689900)(1801.000000,0.691900)(1901.000000,0.698833)(2001.000000,0.704200)(2101.000000,0.712300)(2201.000000,0.711400)(2301.000000,0.714067)(2401.000000,0.732033)(2501.000000,0.730467)(2601.000000,0.728900)(2701.000000,0.742367)(2801.000000,0.730667)(2901.000000,0.737533)
		};
		\addplot[smooth,mycolor3,line width=1pt] plot coordinates{
			(1.000000,0.114900)(101.000000,0.109467)(201.000000,0.112967)(301.000000,0.111200)(401.000000,0.119067)(501.000000,0.155433)(601.000000,0.199400)(701.000000,0.219767)(801.000000,0.242933)(901.000000,0.283500)(1001.000000,0.297833)(1101.000000,0.319133)(1201.000000,0.331100)(1301.000000,0.349567)(1401.000000,0.357600)(1501.000000,0.367267)(1601.000000,0.380467)(1701.000000,0.377200)(1801.000000,0.401900)(1901.000000,0.416767)(2001.000000,0.421067)(2101.000000,0.423233)(2201.000000,0.438467)(2301.000000,0.438700)(2401.000000,0.452233)(2501.000000,0.458600)(2601.000000,0.465167)(2701.000000,0.477800)(2801.000000,0.480433)(2901.000000,0.488067)
		};
		\addplot[densely dashed,mycolor3,line width=1pt] plot coordinates{
			(1.000000,0.257967)(101.000000,0.324867)(201.000000,0.371600)(301.000000,0.396833)(401.000000,0.441200)(501.000000,0.467633)(601.000000,0.482433)(701.000000,0.514833)(801.000000,0.536467)(901.000000,0.572633)(1001.000000,0.593567)(1101.000000,0.613000)(1201.000000,0.638333)(1301.000000,0.650333)(1401.000000,0.666767)(1501.000000,0.680733)(1601.000000,0.693967)(1701.000000,0.702133)(1801.000000,0.713767)(1901.000000,0.721033)(2001.000000,0.724733)(2101.000000,0.737900)(2201.000000,0.739767)(2301.000000,0.746200)(2401.000000,0.748333)(2501.000000,0.756033)(2601.000000,0.758800)(2701.000000,0.763733)(2801.000000,0.765100)(2901.000000,0.776133)
		};
		\end{axis}
		\end{tikzpicture}
		\caption{Orthogonal weights}
		\label{subfig:orth-training-speed-conv-adam}
	\end{subfigure}%
	\caption{ { Learning dynamics of a convolutional ResNet for different initialization scalings \( \sigma_w^2 = 0.01/\sqrt{L} \) (red), \( \sigma_w^2 = 1/\sqrt{L} \) (purple) and \( \sigma_w^2 = 1 \) (blue), with the ReLU nonlinearity. The optimizer is ADAM with an initial  learning rate of \(10^{-2}\). Solid lines without BN and dashed ones with the BN procedure added (for \( \sigma_w^2 = 1 \)). } }
	\label{fig:gauss-orth-training-speed-conv-adam}
\end{figure}

We plot the learning dynamics of the convolutional ResNet-110 for different initialization scalings of random weight with the ReLU nonlinearity in Fig.\ref{fig:gauss-orth-training-speed-conv-adam}. The optimizer is ADAM with an initial learning rate of $10^{-2}.$
\label{sm-subsec:simu-conv}

\end{document}